\documentclass[11pt,a4paper]{article}
\usepackage{times,latexsym}
\usepackage{url}
\usepackage[T1]{fontenc}
\usepackage{longtable}

\usepackage[acceptedWithA]{tacl2018v2}
\usepackage{amsmath}
\usepackage{mathtools}
\usepackage{amssymb}
\usepackage{bbm}
\usepackage{graphicx}
\usepackage{caption}
\usepackage{subcaption}
\usepackage{algorithm}
\usepackage{algpseudocode}
\usepackage{booktabs}
\usepackage{lipsum}
\usepackage{xcolor}
\usepackage{xurl}
\DeclareMathOperator*{\argmax}{arg\,max}

\usepackage{xspace,mfirstuc,tabulary}

\newif\iftaclinstructions
\taclinstructionsfalse 
\iftaclinstructions
\renewcommand{\confidential}{}
\renewcommand{\anonsubtext}{(No author info supplied here, for consistency with
TACL-submission anonymization requirements)}
\newcommand{\instr}
\fi

\iftaclpubformat 

\else

\fi

\newcommand{\com}[1]{}

\usepackage{array}
\newcolumntype{P}[1]{>{\centering\arraybackslash}p{#1}}

\title{Designing an Automatic Agent for Repeated Language based Persuasion Games}

\author{
 Maya Raifer, Guy Rotman, Reut Apel, Moshe Tennenholtz, Roi Reichart \\
  {\sf \{mayatarno, grotman, reutapel\}@campus.technion.ac.il} \\
    {\sf \{roiri, moshet\}@technion.ac.il} \\
}

\date{}

\begin{document}
\maketitle
\begin{abstract}

Persuasion games are fundamental in economics and AI research and serve as the basis for important applications. However, work on this setup assumes communication with stylized messages that do not consist of rich human language. In this paper we consider a repeated sender (expert) -- receiver (decision maker) game, where the sender is fully informed about the state of the world and aims to persuade the receiver to accept a deal by sending one of several possible natural language reviews. We design an automatic expert that plays this repeated game, aiming to achieve the maximal payoff. Our expert is implemented within the Monte Carlo Tree Search (MCTS) algorithm, with deep learning models that exploit behavioral and linguistic signals in order to predict the next action of the decision maker, and the future payoff of the expert given the state of the game and a candidate review. We demonstrate the superiority of our expert over strong baselines and its adaptability to different decision makers and potential proposed deals.\footnote{Our code and data are available at: \url{https://github.com/mayaraifer/automatic_agent}.}

\end{abstract}


\section{Introduction}
\label{sec:introduction}

Natural Language Processing (NLP) has made a substantial progress in recent years, excelling on text understanding applications such as machine translation \cite{bahdanau2014neur,johnson2017google}, information extraction \cite{stanovsky2018supervised} and question answering \cite{andreas2016learning,kwiatkowski2019natural}. However, these applications do not assume that language is used for interaction between strategic participants whose objectives overlap only partially. 

In contrast, in the fields of economics and artificial intelligence (AI), such setups have been widely explored. For example, the settings of personalized advertising and targeted recommendation systems \cite{shapiro1998information,emek2014signaling,bahar2015economic} suggest personalized services for their customers, and solutions are formed as strategic sender-receiver interactions \cite{arieli2019private}. However, these works assume stylized messaging that does not involve real-world natural language.

In this work we address the setting of sender-receiver interaction, but, in contrast to previous research, we assume natural language interaction between the players in an iterative non zero-sum persuasion game. In our setting the two participants are strategic players with their own private utilities. Crucially, the sender has more information about the world than the receiver does. Taking the NLP perspective, we are particularly interested in the persuasion game setting, where the sender's objective is
to persuade the receiver, using natural language messages, to select an action from a set of alternatives. The receiver, in turn, has different payoffs for the different actions. The receiver's payoff depends on properties of the setup that are unavailable to her, and she has a higher level of uncertainty about the setup than the sender has.

Our focus is on repeated non-cooperative setups, where the utilities of the players do not fully overlap. Consider a repeated persuasion game where the interests of the players are aligned. In such a case, the sender should reveal the complete information she posses, letting the receiver take an action which maximizes both their payoffs. In a repeated non-cooperative setup, in contrast, the sender opts to reveal a piece of information that should yield her a high payoff but also maintain a trustful relationship with the receiver, in order to avoid damaging her reputation and hence possibly also her future payoff.


Designing agents to play games is a long standing goal of deep reinforcement learning (RL) research. However, these games are typically zero-sum games, modeled as a utility maximization problem (see e.g., \cite{Silveretal} and the references within). In contrast, in economic contexts like ours, games are rarely zero-sum. A commerce website that aims to recommend a hotel cares about the customer choosing the hotel, while the customer cares about the hotel quality: Their incentives are non-identical, but are also non-opposite. These games cannot be solved as a maximization problem, and there is in fact no optimal player in such problems \cite{fudenberg1991game}. In contrast to economic games where the communication among agents is typically through formal signals or bids \cite{mansour2015bayesian,Baharetal}, we focus on natural language communication which is very natural to persuasion games.

Recently, \citet{apel2020predicting} were the first to adapt the aforementioned setup to natural language messaging. Specifically, they designed a repeated persuasion game in which an \textit{expert} (travel agent) repeatedly interacts with a \textit{decision-maker (DM}, customer). At each trial of the interaction the expert observes a hotel alongside its scored textual reviews, and should choose a single review to reveal to the DM, in a hope to convince her to choose the hotel. The DM, in turn, can choose to either accept or reject the hotel, and her payoff stochastically depends on the review score distribution available to the expert only. Finally, both players observe their payoffs and proceed to the next, similar, step of the game.

While \citet{apel2020predicting} focus on predicting the DM's actions, we adapt their setting and aim to design an \textit{artificial expert (AE)} that should take the expert role in a way that maximizes its payoff. Our AE is implemented within the Monte Carlo Tree Search (MCTS) algorithm, that has been extensively used in AI-based game playing (\S \ref{mcts}). We present language and behavior based deep learning models for two crucial components of the MCTS: (a) A \textit{Decision Making Model (DMM)}, which predicts the actions taken by the DM given the current state of the game; and (b) A \textit{Value Model (VM)}, which predicts the future payoff of the AE given the current state of the game and a potential review that can be presented at the current step. 

We focus on three questions: \textbf{(1)} Can our AE achieve a high payoff? \textbf{(2)} Does our AE adapt its strategy to different decision maker types? and \textbf{(3)} Do our automated AE's strategies resemble those of human AEs ? 

%

%
We test our AE against various types of artificial DMs, compare it to strong alternative experts, and demonstrate its superiority. We further show that our AE is able to adapt its strategy to the DM it faces. We evaluate the impact of proper modeling of the linguistic signal (revealed reviews), comparing a BERT-based approach to hand-crafted features, and show that the later are generally better. Further, we analyze the reviews chosen by our AE, shedding light on its strategy.

Lastly, we also test our AE against human DMs, comparing its performance to a strong baseline. We provide a detailed analysis of the pros and cons of our AE, and discuss the differences between evaluation with human and simulation-based DMs.

%

\section{Related Work}


Some previous works addressed language-based communication in games where the participants have matched or mismatched objectives \cite{golland2010game, frank2012predicting, lewis2017deal}, while other works addressed communication in iterated games \cite{hawkins2017convention}. The main novelty of our setup is the intersection between mismatched objectives and iterative games. We survey relevant works along three lines: Human decision predictions, NLP-based persuasion and artificial agents in textual games.

\paragraph{Human Decision Making Predictions}


Previous work used machine learning to predict human decisions based on non-textual information \cite{altman2006learning,hartford2016deep,plonsky2017psychological}, as well as textual signals, e.g., for judicial decisions \cite{aletras2016predicting,zhong2018legal,medvedeva2020using,yang2019recurrent} and decisions of leading figures \cite{bak2018conversational}. These works formulate the problem as a classification task where the classifier is based on textual (and potentially also other) signals. Unlike in our work, these predictions are not made in a strategic environment, where participants have objectives that affect their decisions.

Several works aim to draw predictions of human decisions in competitive games given textual signals \cite{ben2020predicting, oved2020predicting}. For example, \citet{niculae2015linguistic} proposed an algorithm for predicting actions in an online strategy game based on the language produced by the players as part of the inter-player communication required in the game. The setups of these works differ from ours, and, particularly, they do not address persuasion and repeated games.


The most relevant work to ours is that of \citet{apel2020predicting}: We use their setup and data (\S \ref{Task Definition}). However, \citet{apel2020predicting} only focused on predicting the decisions of the decision-maker. In addition, while they based their predictions on past and future game information, we perform more realistic predictions based on past information only.

\paragraph{Persuasion in NLP}

\citet{hidey2017analyzing} proposed an annotation scheme to differentiate claims and premises using different persuasion strategies in an online persuasive forum \cite{tan2016winning}. \citet{hidey2018persuasive} tried to predict persuasiveness in social media posts containing sequential arguments. \citet{yang2019let}, \citet{wang2019persuasion} and \citet{chen2021weakly} aimed to quantify persuasiveness and to identify persuasive strategies. This line of works, which aims to analyze and predict persuasive aspects of language, is a step towards developing persuasive agents.
 
Several works studied persuasion dialogue tasks. While models for task oriented dialogue have achieved promising performance on tasks where the users and the system are coordinated in their goals, persuasion dialogue tasks are less common. \citet{hiraoka2014reinforcement} focused on learning a policy which satisfies both user and system goals in a cooperative persuasive dialogue. \citet{li2020end} proposed an end-to-end neural network to generate diverse coherent responses for non-collaborative dialogue tasks, where users and systems do not share a common goal. \citet{efstathiou2014learning} developed a dialogue agent which learns to perform non-cooperative dialogue turns for utility maximization in a stochastic trading game with very simple linguistic messages. \citet{lewis2017deal} trained end-to-end models for negotiation in a semi-cooperative setup. These works differ from ours since we focus on designing an artificial agent in a repeated persuasion game setting, where the expert should construct a long term strategy as its choice in a specific trial affects both the outcome of that trial and its future reputation.


\paragraph{Artificial Agents In Textual Games}

Several works designed agents for referential games \cite{lazaridou2016multi,havrylov2017emergence}, where agents should interactively develop a shared language in order to communicate with each other and solve a joint task. Another line of work designs agents for games inspired by \citet{wittgenstein1953philosophical}'s language games \cite{wang2016learning}, where a human aims to accomplish a task (e.g., achieving a certain configuration of blocks), but is only able to communicate with an artificial agent which performs the actual actions. Such games are cooperative in nature as the players share their goals. Finally, \citet{narasimhan2015language} address text-based games, where natural language is used both to describe the state of the world and the actions of the participating players. They design a deep RL agent that jointly learns state representations and action policies using game rewards as feedback. This game is also very different from ours.

\section{Task Definition}\label{Task Definition}

We consider a two-player, travel agent (expert) and customer (decision-maker, DM), repeated persuasion game . The game, first introduced by \citet{apel2020predicting}, consists of a sequence of ten trials. In each trial, the expert observes seven reviews of a given hotel, alongside their scores, and she then sends the DM one of the reviews, without its score. Based on this review, the DM decides between two options: Accepting or rejecting the hotel.
If the hotel is not accepted by the DM, the payoff of both players is 0. Otherwise, the expert's payoff is 1, and the DM's payoff is a score randomly sampled from the seven scores presented to the expert at the beginning of this trial, referred to as \textit{the lottery result}, minus the constant 8. This constant imposes a zero expected payoff for a DM who chooses to accept the hotel in all the ten trials.\footnote{For full information of the train and test hotels, including their review scores, see Table 1 of \citet{apel2020predicting}.}

A more abstract description of each trial in this multi-stage game would be as follows. Every hotel is associated with an unknown distribution over payoffs, corresponding to the distribution over experiences that guests will have at this hotel. The scored reviews are sampled from this distribution, and the DM's reward is another sample from the distribution. Since in our setting we do not have access to the real payoff distribution of each hotel, we approximate it using the empirical distribution from the payoffs observed by the expert.

Formally, denote the suggested hotel at trial \(t\) by \(h_t\), the DM's decision at this trial by \(a_t\), where \(a_t=1\) if the DM accepts the hotel, and the seven scores attached to the reviews of \(h_t\) by \(s^t_{1},s^t_{2},..s^t_{7}\), where $s^t_{i} \in [0,10]$. The players' payoffs are: 
\begin{equation*}
\textit{expert-payoff} = \mathbbm{1}_{\{a_t=1\}},
\end{equation*}
\begin{equation*}
\begin{split}
\textit{dm-payoff} = \mathbbm{1}_{\{a_t=1\}}\cdot (s^t_{i}-8), \\ i \sim uniform[1,7]. \hspace{1.5cm}\end{split}
\end{equation*}
While the two players would ideally like to gain the highest possible payoff (i.e., this is not a zero-sum game), their strategies are not necessarily coordinated. Particularly, while the expert aims to sell as many hotels as possible, the DM aims to accept only hotels which are likely to yield a positive payoff. Note that the DM is not fully informed of the hotel state, and should make her decision based on the partial information provided by the expert. The repeated nature of the game adds complexity to the decisions, as the expert's choice in a specific trial affects not only the DM's decision in this trial but also the expert's reputation in the next trials. 
 
 Let us consider the game from the expert's point of view. Consider an expert who cares solely about the present and reveals a high-score review in order to tempt the DM to choose the hotel, even if the acceptance decision is likely to yield a negative payoff. This expert is likely to gain a high payoff at the first few rounds. However, as the game proceeds the DM would probably understand that the expert is unreliable. On the other hand, if the expert reveals only reviews that reliably describe the hotel (e.g., the median scoring reviews), the DM is likely not to choose the hotel when she is presented with mediocre reviews.
 
\citet{apel2020predicting} provide an equilibrium analysis of our game. This is a theoretical analysis, under some constraining assumptions and as the authors demonstrate the players do not follow it in practice. This further motivates our work which aims to design an NLP-based agent of the expert in this game. Note, that our approach is different from that of \citet{apel2020predicting} who aimed to predict individual decisions of the DM, rather than constructing an artificial DM or expert.


\paragraph{Data}\label{data}

We use the dataset collected by \newcite{apel2020predicting} using Amazon Mechanical Turk. \footnote{\url{https://github.com/reutapel/Predicting-Decisions-in-Language-Based-Persuasion-Games}} The dataset is composed of 509 ten-trial games. The participants were randomly and anonymously paired, and each of them was randomly selected to be in one of the two roles: DM or expert. 

The training set consists of 408 games. In these games the same hotels and reviews were used, but the hotels were randomly permuted between the 10 trials. The test set consists of 101 games, played with a different set of hotels and reviews, such that the hotels are again randomly permuted. Each participant was allowed to participate in the experiment only once, such that the training and test sets consist of different players.

\begin{figure}[!t]
  \centering
    \includegraphics[width=0.5\textwidth]{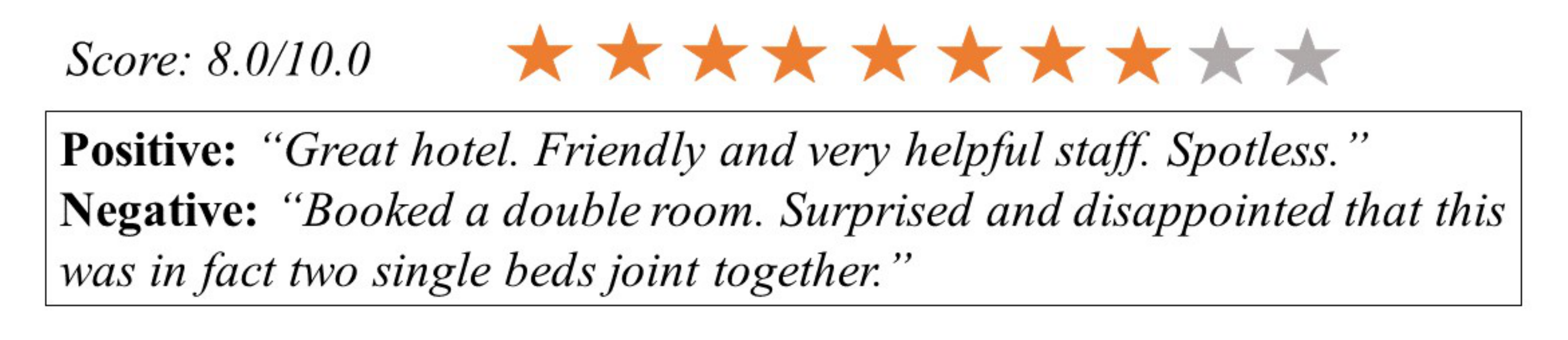}
  \caption{An example review from the \citep{apel2020predicting}'s dataset. Each review consists of a continuous score ranging from 0 to 10, alongside positive and negative textual descriptions.}
\label{fig:example_review}
\end{figure}

Each hotel is accompanied with seven reviews collected from the Booking.com website along with their scores, continuously ranging between 0 and 10 (see an example review in Figure \ref{fig:example_review}).
All the reviews contain at least 100 characters and are separated into positive and negative parts. Figure \ref{fig:example_review} demonstrates a sampled review from the dataset. The order in which each of these parts were presented to the experts was also assigned at random. For more details see \citet{apel2020predicting}.

\section{Method}

We design an AE which aims to maximize its payoff in the persuasion game.

\paragraph{The High-level Structure of our Algorithm}

Our algorithm is composed of three  components:\\
\textbf{(a) MCTS} -- an online search algorithm which looks for the best action out of a predefined set (in terms of maximum expected payoff) at each game trial. In our setting, actions correspond to review selection, so the MCTS determines which review should be revealed to the DM in each trial.\newline
\textbf{(b) The DM Model (DMM)} -- a model which predicts the decision made by the DM in each trial of the game. This model allows the MCTS algorithm to simulate the DM's response to revealed reviews.\newline
\textbf{(c) The Value Model (VM)} -- a model which predicts the expert's future payoff in each trial of the game. It is used by the MCTS to initialize the expected return values of new explored decision paths.

Note that MCTS is the core component of our AE and the two other models are integrated into it after they have been trained offline.  We next describe these three components in detail, concluding the section with a description of the two feature sets used by the DMM and the VM.

\subsection{The MCTS Algorithm}
\label{mcts}

MCTS \cite{coulom2006efficient} is a heuristic search technique, presented in the field of RL. It has received considerable attention due to its success in the difficult problem of computer Go \cite{gelly2006mogo} and has been used widely in challenging domains such as general game playing \cite{finnsson2008simulation,kim2017opponent,baier2018evolutionary,sironi2018self} and real-time strategy games \cite{balla2009uct,ontanon2016informed}. We briefly describe MCTS in the context of our game settings. A detailed survey can be found at \citet{coulom2006efficient} and \citet{browne2012survey}.

The MCTS determines the best action out of a set of available actions by balancing the exploration-exploitation trade-off.
It constructs a search tree, node-by-node, starting from a root node defined by the current state of the game.
In our setting, \(s(v)\), the state of the node \(v\), is uniquely defined by the complete history of the game and the current suggested hotel \(h\). Therefore, the action space \(A(s(v))\) of \(s(v)\) consists of the corresponding reviews of its current suggested hotel \(h\), \(A(s(v))=\{r_{hi}| i \in \{1,..7\}\}\), where \(r_{hi}\) denotes the \(i'th\) review of hotel \(h\).

We initialize the values of each state node variable \(s(v)\) according to our VM function, to predict its expected future payoff. For each trial $t$ of the game the MCTS is provided with the new candidate hotel, and the next steps of the game are simulated with the VM and DMM. Based on this simulation the algorithm selects the optimal expert action, i.e., the optimal review that should be revealed to the DM. 

\com{
\subsection{The MCTS Algorithm}\label{mcts}

MCTS \cite{coulom2006efficient}, Algorithm \ref{pseudocode}) is a heuristic search technique, presented in the field of RL. It has received considerable attention due to its success in the difficult problem of computer Go \cite{gelly2006mogo} and has been used widely in challenging domains such as general game playing \cite{finnsson2008simulation,kim2017opponent,baier2018evolutionary,sironi2018self} and real-time strategy games \cite{balla2009uct,ontanon2016informed}.


The MCTS determines the best action out of a set of available actions by balancing the exploration-exploitation trade-off.
It constructs a search tree, node-by-node, starting from a root node defined by the current state of the game.
In our setting, \(s(v)\), the state of the node \(v\), is uniquely defined by the complete history of the game and the current suggested hotel \(h\). Therefore, the action space \(A(s(v))\) of \(s(v)\) consists of the corresponding reviews of its current suggested hotel \(h\), \(A(s(v))=\{r_{hi}| i \in \{1,..7\}\}\), where \(r_{hi}\) denotes the \(i'th\) review of hotel \(h\).
In addition, each node \(v\) holds two values, updated during the search process. The first is \(Q(s(v),r)\) \( \forall r \in A(s(v))\), representing the expected return after review \(r \in A(s(v))\) is chosen in state \(s(v)\). The second is \(N(s(v),r)\), counting the number of times that review \(r\) has been selected in \(s(v)\). \par
The algorithm can be broken down into four modules: selection, expansion, simulation and backpropagation.

\textbf{(a)} Selection -- the MCTS algorithm scans the current tree from the root to a leaf node using a specific strategy. The common strategy is the Upper Confidence Bound for Trees (UCT), which balances the exploration-exploitation trade-off: 
\begin{equation*}
\begin{split}
\pi_{UCT}(s(v)) = \argmax_{r\in A(s(v))} Q(s(v),r) +\\ c\sqrt{\frac{N(s(v),r)}{\sum_{r \in A(S(v))}N(s(v),r)}}, \end{split}
\end{equation*}
where $c$ is an exploration parameter.

\textbf{(b)} Expansion -- a new child node is added to the tree as a leaf node after it was reached during the selection process for the first time. 
While in some works \(Q(s(v),r)\) of a new node is estimated only by simulation, we train offline the VM and initialize \(Q(s(v),r)\) according to its prediction.

\textbf{(c)} Simulation -- Once reaching a new child node, a simulation is performed by choosing reviews according to a random policy until reaching a terminal node. The DM's response to this review is simulated using the offline trained DMM.

\textbf{(d)} Backpropagation -- After observing the total payoff $p$ at the end of the simulation, we update the values of all observed nodes along the chosen path. That is, for each review \(r\) chosen in state \(s(v)\) during the current simulation, we have:
\begin{equation*}
    \begin{split}
      N(s(v),r) \leftarrow N(s(v),r) + 1 \\
      Q(s(v),r) \leftarrow Q(s(v),r) \hspace{0.15cm} + \hspace{0.35cm}\\
      \frac{1}{N(s(v),r)}\cdot(p-Q(s(v),r))  
    \end{split}
\end{equation*}

\begin{algorithm}
	\caption{MCTS with UCT} 
	\label{pseudocode}
	\begin{algorithmic}
	\State\textbf{Input: \(v_0\) root state}
	\State \textbf{Output:} best possible review $r^*$
	\While{within time limit}
			 \State $v_s \leftarrow \sc{selection}(v_0)$
			 \State $v_l \leftarrow \sc{expansion} (v_s)$
			\State $p \leftarrow \sc{simulation}(s(v_l))$
            \State $\sc{backpropagation(v_l,p)}$
    \EndWhile 
          \State\textbf{return} $r^* = \argmax_{r \in A(s(v_0))}Q(s(v_0),r)$
	\end{algorithmic} 

	\end{algorithm}
}

\subsection{The DMM \& VM Models}\label{sec:DMM-VM}

\com{
\begin{figure}[!h]
\centering
  \centering
  \begin{subfigure}[b]{0.2\textwidth}
    \includegraphics[\textwidth,height=4cm]{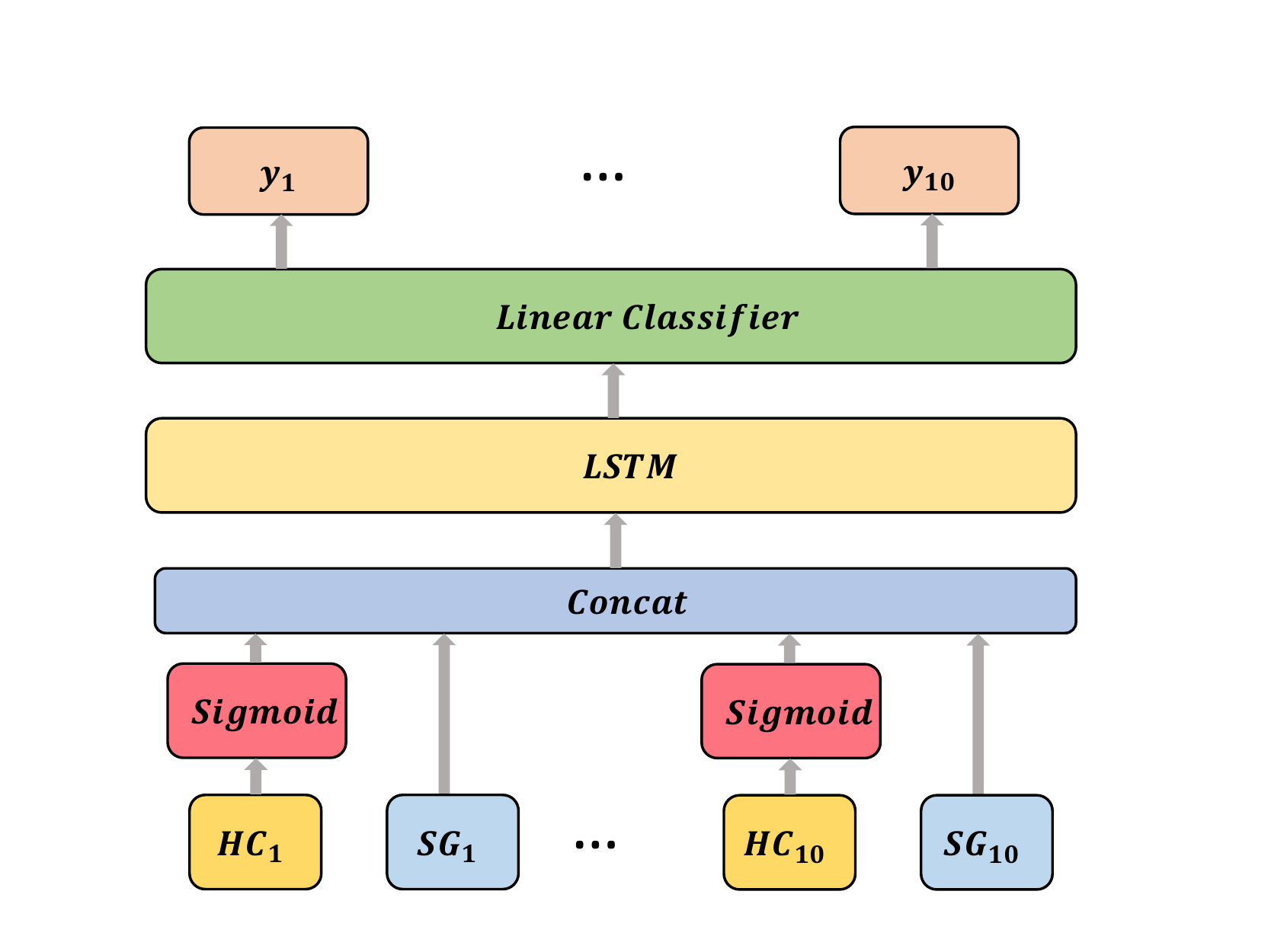}
    \caption{HC-LSTM}
  \end{subfigure}
  \hfill
  \begin{subfigure}[b]{0.22\textwidth}
    \includegraphics[\textwidth,height=4cm]{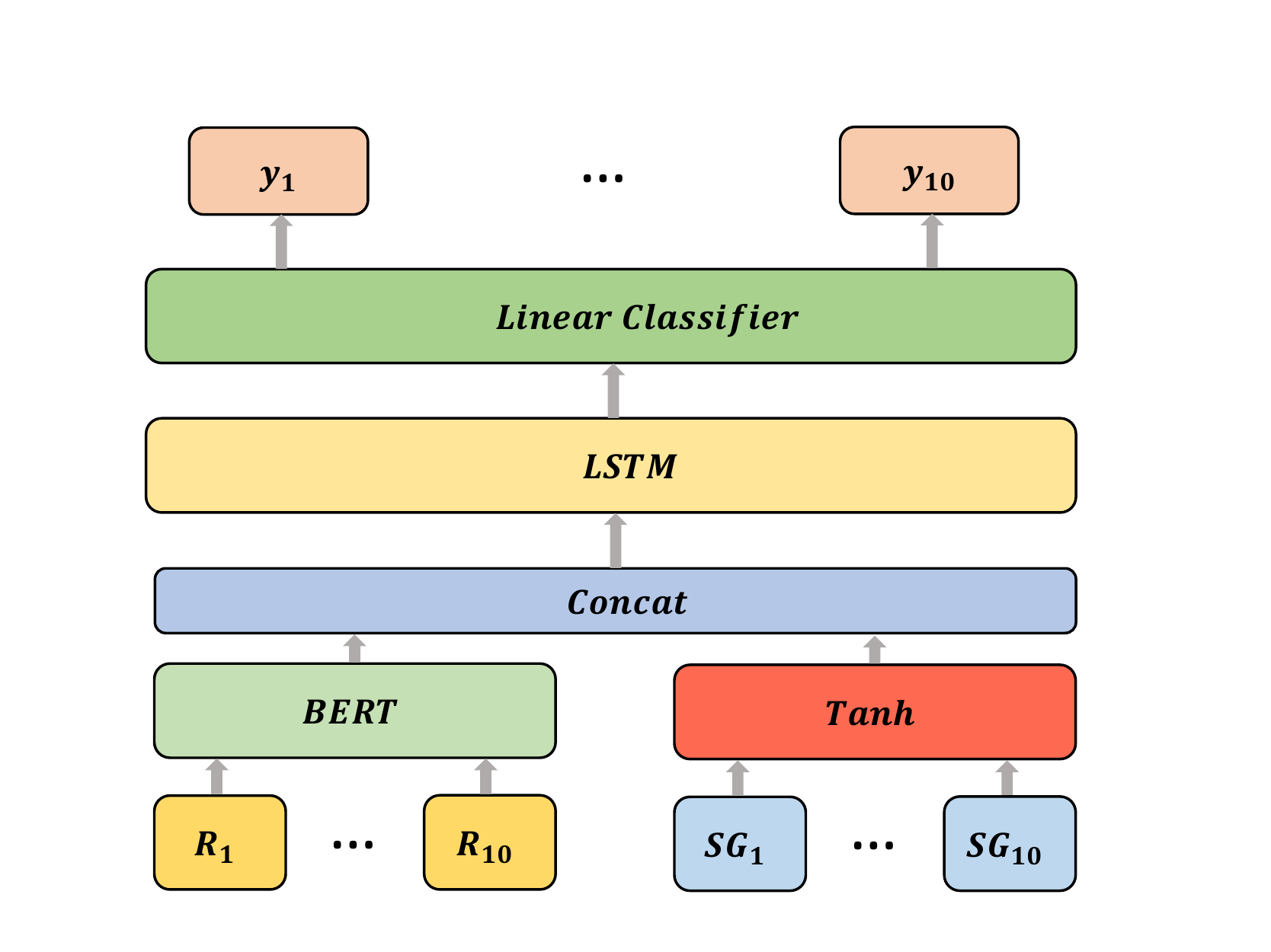}
    \caption{BERT-LSTM}
  \end{subfigure}
  \caption{Illustration of our two model architectures. 
  \(HC_t\), \(SG_t\) and \(R_t\) denote the hand-crafted features, the statistical game features and the presented review in trial $t$, respectively. For DMM, \(y_t\) is the DM's decision in trial \(t\), and for VM, \(y_t\) is the expert's future payoff in trial \(t\). 
  }
\label{fig:model_architecture}
\end{figure}
}

\begin{figure*}[!ht]
    \centering
    \subfloat[\centering HC-LSTM]
    {{\includegraphics[height=5.2cm,width=0.4\textwidth]{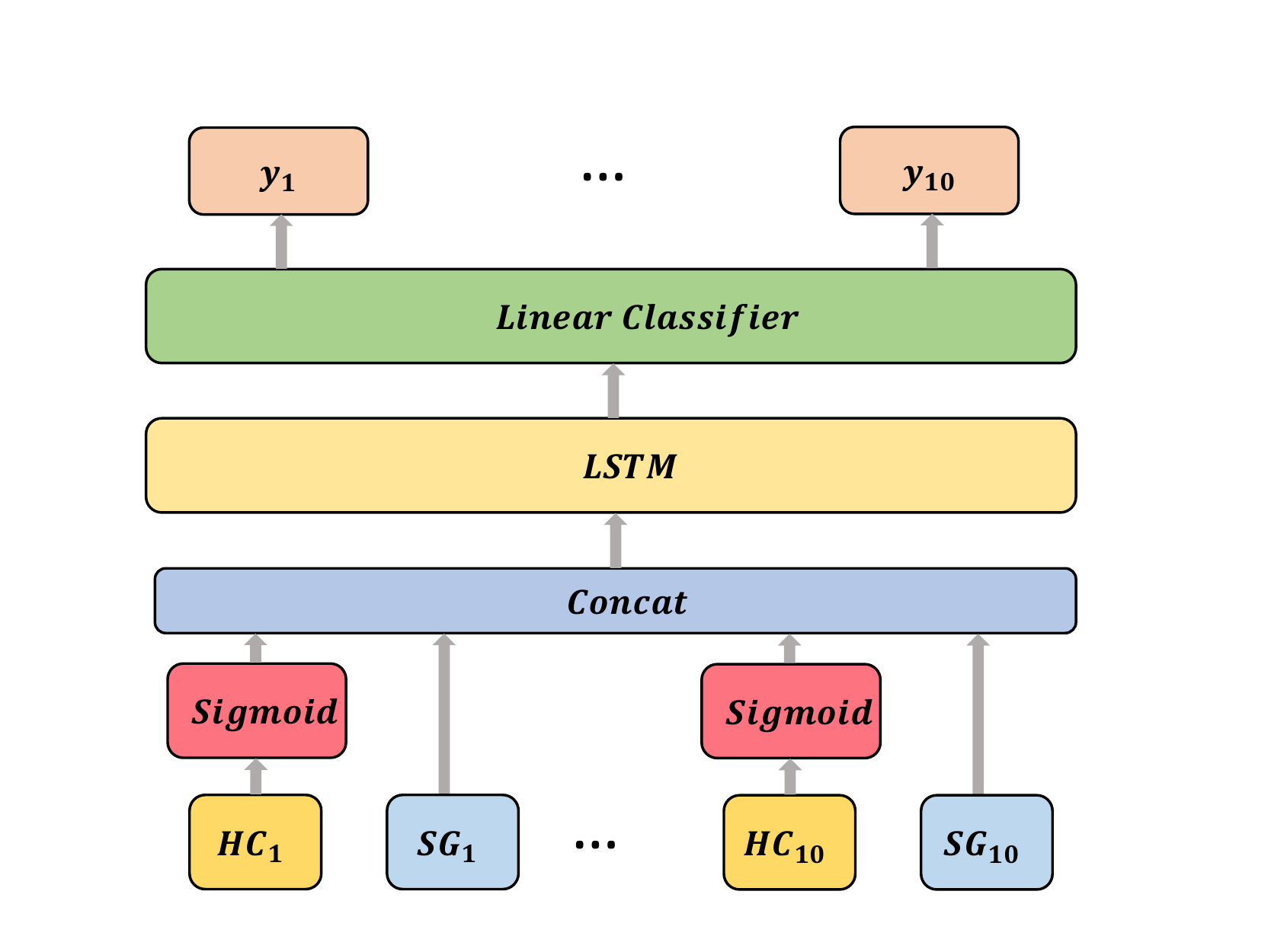}}}
    \qquad
    \centering
    \subfloat[\centering BERT-LSTM]
    {{\includegraphics[height=5.2cm,width=0.4\textwidth]{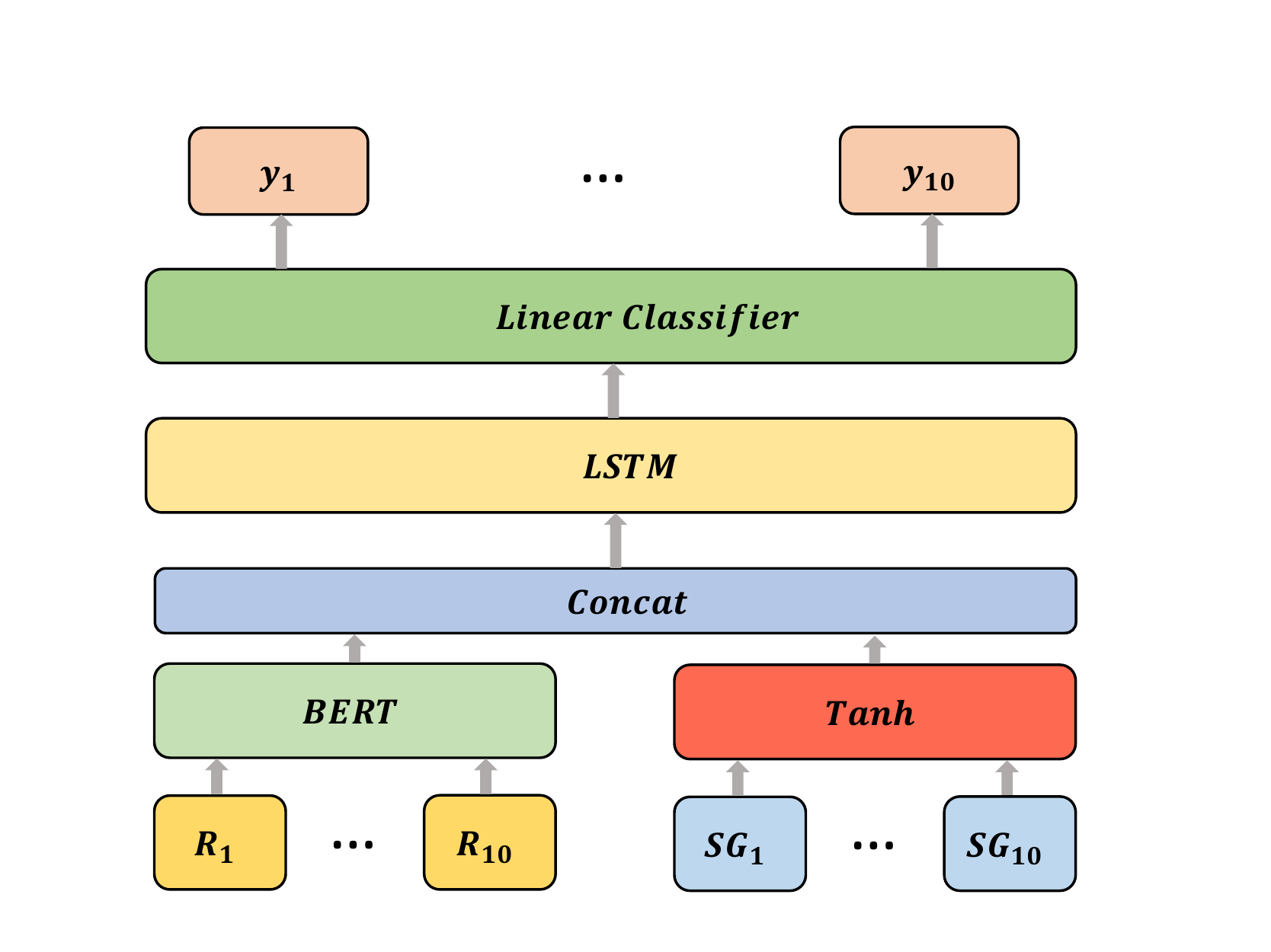}}}
    \caption{Illustration of our two model architectures. 
  \(HC_t\), \(SG_t\) and \(R_t\) denote the hand-crafted features, the statistical game features and the presented review in trial $t$, respectively. For DMM, \(y_t\) is the DM's decision in trial \(t\), and for VM, \(y_t\) is the expert's future payoff in trial \(t\).} 
    \label{fig:model_architecture}
\end{figure*}


The DMM and the VM are applied in each trial of the game, for predicting the DM's decision (DMM) and the expert's future payoff (VM).
%
The predictions at trial $t$ are based on information about the previous trials and the current trial. Both models have identical architectures, and they are trained off-policy on the training set of \citet{apel2020predicting}. Due to the different nature of prediction, however, they are trained to optimize different loss functions: Binary cross entropy (DMM) and mean squared error (VM). In both cases training is done with the Adagrad algorithm \cite{duchi2011adaptive}.

We consider two architectures (Figure \ref{fig:model_architecture}). Due to the sequential nature of the decision making process, we based the two models on the Long Short-Term Memory (LSTM) architecture \cite{hochreiter1997long}.
We feed the first LSTM variant, denoted by HC-LSTM, with two types of features: (a) statistical game features, representing the information about the previous and the current trials; and (b) hand-crafted textual features \cite{apel2020predicting}, automatically extracted from the review. A detailed description of both types of features is provided in \S \ref{features}. The binary hand-crafted features are passed through the Sigmoid activation function and are concatenated to the continuous statistical game features before being passed to the LSTM encoder.

The second architecture, denoted by BERT-LSTM, is an LSTM fed by the statistical game features and the pooler output of BERT \cite{devlin2018bert}. Since the encoded output of BERT is processed by the Tanh activation function, we pass the statistical game features through it before performing the concatenation and passing the resulted vectors to the LSTM encoder.

\subsection{Features}\label{features}

We explore two types of hand-crafted features: Hand-crafted textual features (HC), capturing textual knowledge from the reviews, and statistical game features (SG), capturing properties of the human interactions during the game. 

The HC set, consisting of 42 binary features that can be split into three feature types, was created by \citet{apel2020predicting}. Features of the first type indicate whether some predefined topics are mentioned in the positive and negative parts of the review (e.g., facilities, price, location, staff, transportation, food, etc.). Features of the second type correspond to predefined textual properties of the positive and negative parts of the review, e.g., the length of each part (short/medium/long), existence of words with high, medium or low intensity, etc. Finally, features of the third type capture the structural properties of the overall review, e.g., the ratio between the lengths of the positive and negative parts. While these features are hand-crafted, they are automatically extracted from the text. We refer the reader to \citet{apel2020predicting} for further details.

Table \ref{featurestable} provides a detailed description of the SG features, some of which are a contribution of this paper. The SG set includes two main types of features: (a) Features that represent information about the DM's behavior up to trial \(t\). For example, HotelAcceptance measures the proportion of trials where the DM accepted a hotel; and (b) Features that represent general information about the game up to trial $t$. For example, the proportion of trials where the lottery result was low, high or medium and whether the proposed hotel has a low, high or medium average score.

\com{
\begin{table*}[!h]
\centering
\resizebox{0.88\textwidth}{7.5cm}{
\begin{tabular*}{\textwidth}{
|p{3cm}|p{8.2cm}|p{3.5cm}|}
\hline
\textbf{Feature Name} & \textbf{Feature Description} & \textbf{Feature Formulation } \\
\hline
\hline

\multicolumn{3}{|l|} {\textbf{Behavioral Features}}\\
\hline
\hline
HotelAcceptance &Avg \#trials where the hotel was accepted & \(\frac{\sum_{i=1}^{t-1}         \mathbbm{1}_{\{a_i=1\}}}{ t-1}\) \\ 
\hline
HotelAcceptance Earn & Avg \#trials where the hotel was accepted and the DM achieved a negative payoff.*& \(\frac{\sum_{i=1}^{t-1} \mathbbm{1}_{\{a_i=1 \cap dmp_i>0\}}}{t-1}\) \\
\hline
HotelAcceptance Lose&Avg \#trials where the hotel was accepted and the DM achieved a positive payoff.*&\(\frac{\sum_{i=1}^{t-1} \mathbbm{1}_{\{a_i=1 \cap dmp_i<0\}}}{ t-1}\)\\
\hline
\(\neg\)HotelAcceptance Earn&Avg \#trials where the hotel was not accepted but the payoff would have been positive if the DM had accepted it.*&\(\frac{\sum_{i=1}^{t-1} \mathbbm{1}_{\{a_i=0 \cap dmp_i>0\}}}{ t-1}\) \\
\hline
 \(\neg\)HotelAcceptance Lose&Avg \#trials where the hotel was not accepted but the payoff would have been negative if the DM had accepted it.*& \(\frac{\sum_{i=1}^{t-1} \mathbbm{1}_{\{a_i=0 \cap dmp_i<0\}}}{ t-1}\) \\ 
\hline
 BadHotel \hspace{1.5cm}Acceptance & Avg \#trials where a hotel with average score lower than 7.5 was accepted. & \(\frac{\sum_{i=1}^{t-1} \mathbbm{1}_{\{a_i=1 \cap s(hi)<7.5\}}}{ t-1}\) \\ 
    \hline
  \(\neg\)ExcellentHotel Acceptance &Avg \#trials where a hotel with average score higher than 9.5 was accepted.& \(\frac{\sum_{i=1}^{t-1} \mathbbm{1}_{\{a_i=0 \cap s(hi)>9.5\}}}{ t-1}\) \\ 
   \hline
   DMPayoff &Avg DM's payoff pet trial& \(\frac{\sum_{i=1}^{t-1} dmp_i}{t-1}\) \\
  \hline
   \multicolumn{3}{ |l| }{\textbf{General Features}} \\

  \hline
   LotteryLow & Avg \#trials where the lottery result was lower than 3.*& \(\frac{\sum_{i=1}^{t-1} \mathbbm{1}_{\{l_i<3\}}}{t-1}\) \\ 
 \hline
 LotteryMed &Avg \#trials where the lottery result was between 3 to 5.*& \(\frac{\sum_{i=1}^{t-1} \mathbbm{1}_{\{l_i\geq3 \cap l_i<5\}}}{t-1}\) \\ 
  \hline
  LotteryHigh &Avg \#trials where the lottery result was higher than 8.*&  \(\frac{\sum_{i=1}^{t-1} \mathbbm{1}_{\{l_i\geq8\}}}{t-1}\) \\ 
  \hline
   CompletedTrials & The proportion of trials that have already been played. & \(\frac{{t-1}}{10}\) \\ 
  \hline
  GoodHotel &Avg score of the current hotel is higher than 8.5. & \(\mathbbm{1}_{\{s(h_t)\geq8.5\}}\) \\ 
 \hline
  MedHotel &Avg score of the current hotel is between 7.8 to 8.5.& \(\mathbbm{1}_{\{s(h_t)<8.5 \cap s(h_t)\geq 7.5\}}\) \\ 
 \hline
  BadHotel &Avg score of the hotel is lower than 7.5.& \(\mathbbm{1}_{\{s(h_t)\leq7.5\}}\) \\  
 \hline
  HighScore& The attached score of the presented review is higher than 8.5.& \(\mathbbm{1}_{\{s(r_t)\geq8.5\}}\) \\  
  \hline
 MedScore &The attached score of the presented review is between 7.5 to 8.5. & \(\mathbbm{1}_{\{s(r_t)<8.5 \cap s(r_t)\geq7.5\}}\) \\  
 \hline
  LowScore &The attached score of the presented review is lower than 7.5.& \(\mathbbm{1}_{\{s(r_t)<7.5\}}\) \\  
  \hline
  TopReview &The attached score of the presented review is in the top 3 scoring reviews.& \(\mathbbm{1}_{\{s(r_t) \in\text{ top 3 scores}\}}\) \\  
  \hline
  BottomReview &The attached score of the presented review is not in the top 3 scoring reviews.&  \(\mathbbm{1}_{\{s(r_t)\notin\text{ top 3 scores}\}}\) \\  
 \hline

\end{tabular*}}
\caption{SG features of trial $t$. \(a_{i}\), \(l_i\) and \(dmp_{i}\) denote the DM's action, lottery result and DM's payoff in trial $t$, respectively.
\(s(h_t)\) is the average score of the suggested hotel in trial $t$, \(r_t\) is its revealed review and \(s(r_t)\) is the revealed review score. * indicates that the feature is taken from \citet{apel2020predicting}.}
\label{featurestable}
\end{table*}}

\begin{table*}[!h]
\centering
\scalebox{0.75}{
\begin{tabular*}{\textwidth}{
|p{3cm}|p{8.2cm}|p{3.5cm}|}
\hline
\textbf{Feature Name} & \textbf{Feature Description} & \textbf{Feature Formulation } \\
\hline
\hline

\multicolumn{3}{|l|} {\textbf{Behavioral Features}}\\
\hline
\hline
HotelAcceptance &Avg \#trials where the hotel was accepted & \(\frac{\sum_{i=1}^{t-1}         \mathbbm{1}_{\{a_i=1\}}}{ t-1}\) \\ 
\hline
HotelAcceptance Earn & Avg \#trials where the hotel was accepted and the DM achieved a negative payoff.*& \(\frac{\sum_{i=1}^{t-1} \mathbbm{1}_{\{a_i=1 \cap dmp_i>0\}}}{t-1}\) \\
\hline
HotelAcceptance Lose&Avg \#trials where the hotel was accepted and the DM achieved a positive payoff.*&\(\frac{\sum_{i=1}^{t-1} \mathbbm{1}_{\{a_i=1 \cap dmp_i<0\}}}{ t-1}\)\\
\hline
\(\neg\)HotelAcceptance Earn&Avg \#trials where the hotel was not accepted but the payoff would have been positive if the DM had accepted it.*&\(\frac{\sum_{i=1}^{t-1} \mathbbm{1}_{\{a_i=0 \cap dmp_i>0\}}}{ t-1}\) \\
\hline
 \(\neg\)HotelAcceptance Lose&Avg \#trials where the hotel was not accepted but the payoff would have been negative if the DM had accepted it.*& \(\frac{\sum_{i=1}^{t-1} \mathbbm{1}_{\{a_i=0 \cap dmp_i<0\}}}{ t-1}\) \\ 
\hline
 BadHotel \hspace{1.5cm}Acceptance & Avg \#trials where a hotel with average score lower than 7.5 was accepted. & \(\frac{\sum_{i=1}^{t-1} \mathbbm{1}_{\{a_i=1 \cap s(hi)<7.5\}}}{ t-1}\) \\ 
    \hline
  \(\neg\)ExcellentHotel Acceptance &Avg \#trials where a hotel with average score higher than 9.5 was accepted.& \(\frac{\sum_{i=1}^{t-1} \mathbbm{1}_{\{a_i=0 \cap s(hi)>9.5\}}}{ t-1}\) \\ 
   \hline
   DMPayoff &Avg DM's payoff pet trial& \(\frac{\sum_{i=1}^{t-1} dmp_i}{t-1}\) \\
  \hline
   \multicolumn{3}{ |l| }{\textbf{General Features}} \\

  \hline
   LotteryLow & Avg \#trials where the lottery result was lower than 3.*& \(\frac{\sum_{i=1}^{t-1} \mathbbm{1}_{\{l_i<3\}}}{t-1}\) \\ 
 \hline
 LotteryMed &Avg \#trials where the lottery result was between 3 to 5.*& \(\frac{\sum_{i=1}^{t-1} \mathbbm{1}_{\{l_i\geq3 \cap l_i<5\}}}{t-1}\) \\ 
  \hline
  LotteryHigh &Avg \#trials where the lottery result was higher than 8.*&  \(\frac{\sum_{i=1}^{t-1} \mathbbm{1}_{\{l_i\geq8\}}}{t-1}\) \\ 
  \hline
   CompletedTrials & The proportion of trials that have already been played. & \(\frac{{t-1}}{10}\) \\ 
  \hline
  GoodHotel &Avg score of the current hotel is higher than 8.5. & \(\mathbbm{1}_{\{s(h_t)\geq8.5\}}\) \\ 
 \hline
  MedHotel &Avg score of the current hotel is between 7.8 to 8.5.& \(\mathbbm{1}_{\{s(h_t)<8.5 \cap s(h_t)\geq 7.5\}}\) \\ 
 \hline
  BadHotel &Avg score of the hotel is lower than 7.5.& \(\mathbbm{1}_{\{s(h_t)\leq7.5\}}\) \\  
 \hline
  HighScore& The attached score of the presented review is higher than 8.5.& \(\mathbbm{1}_{\{s(r_t)\geq8.5\}}\) \\  
  \hline
 MedScore &The attached score of the presented review is between 7.5 to 8.5. & \(\mathbbm{1}_{\{s(r_t)<8.5 \cap s(r_t)\geq7.5\}}\) \\  
 \hline
  LowScore &The attached score of the presented review is lower than 7.5.& \(\mathbbm{1}_{\{s(r_t)<7.5\}}\) \\  
  \hline
  TopReview &The attached score of the presented review is in the top 3 scoring reviews.& \(\mathbbm{1}_{\{s(r_t) \in\text{ top 3 scores}\}}\) \\  
  \hline
  BottomReview &The attached score of the presented review is not in the top 3 scoring reviews.&  \(\mathbbm{1}_{\{s(r_t)\notin\text{ top 3 scores}\}}\) \\  
 \hline

\end{tabular*}}
\caption{SG features of trial $t$. \(a_{i}\), \(l_i\) and \(dmp_{i}\) denote the DM's action, lottery result and DM's payoff in trial $t$, respectively.
\(s(h_t)\) is the average score of the suggested hotel in trial $t$, \(r_t\) is its revealed review and \(s(r_t)\) is the revealed review score. * indicates that the feature is taken from \citet{apel2020predicting}.}
\label{featurestable}
\end{table*}

\section{Experiments}
\label{sec:experiments}

\paragraph {Experimental Setting}
\label{ArtificialExpertExperimentalSetting}

Evaluating our AE against humans is highly expensive and time consuming, and hence infeasible at large scales. We hence start with another, widely used solution: Human simulations \cite{jung2008integrated,ai2008user,gonzalez2010cooperative,shi2019build,zhang2020evaluating}. In this approach we evaluate the AE against an automatic algorithm that simulates human DMs. While this evaluation is not performed against actual humans, it allows us to evaluate the AE against various types of players, by changing the data-driven DM in a controlled manner. We perform 1000 simulated games over the test set per DM simulator, where the order in which the hotels are presented to the AE is randomly permuted at each simulation. 

We employ two DMMs (\textbf{HC-LSTM} and \textbf{BERT-LSTM}) as our basic DM simulators, as they are trained to imitate the human DM's behavior in the game. We further modify the behavior of these "human like" DMMs, by changing their hotel acceptance probability in a controlled manner. We consider: \textbf{(a)} \textbf{\(\alpha\)-compromised} DMMs, where the acceptance probability is increased by $\alpha = 0.1$ or $\alpha = 0.2$ over the prediction of the basic DMM; and \textbf{(b)} \textbf{\(\alpha\)-inflexible} DMMs, where the acceptance probability is similarly decreased.


\paragraph{Baselines}

We next describe the baselines for the AE and for its components, the DMM variants (\textbf{HC-LSTM} and \textbf{BERT-LSTM}), and the VM variants (\textbf{HC-LSTM} and \textbf{BERT-LSTM}).

\textbf{\textit{DMM.}} The DMM decides in each trial whether to accept a suggested hotel or not.
We propose four different DMM variants, differing in their decision strategy, architecture and features:
\textbf{(a)} \textbf{HC-SVM} -- a Support Vector Machine (SVM, \cite{cortes1995support}) based on the HC and SG features. It allows us to evaluate the power of a non-DNN and non-sequential modeling approach;
\textbf{(b)} \textbf{BERT-SVM} -- This model is similar to HC-SVM, except that the text is represented with BERT;
\textbf{(c)} \textbf{Expected Weighted Guess (EWG)} -- a random baseline which applies the hotel acceptance probability of the training set ($p = 0.72$); and
\textbf{(d)} \textbf{Previous Decisions (PD)} -- a deterministic baseline which predicts that the DM accepts the hotel only if it accepted at least half of the previous hotels.

\textbf{\textit{VM.}} The VM predicts the expert's future payoff in each trial. We propose five different variants of it:
\textbf{(a)} \textbf{HC-SVR} -- a Support Vector Regression (SVR) \cite{drucker1997support} model based on the HC and SG features. This is a non-DNN and non-sequential approach;
\textbf{(b)} \textbf{BERT-SVR} -- an SVR model based on the SG and the encoded BERT features;
\textbf{(c)} \textbf{Maximal Future Payoff (MFO)} -- a deterministic baseline that assumes that all future hotels will be accepted and hence the future payoff at each trial is maximal;
\textbf{(d)} \textbf{Average Value (AV)} -- a deterministic baseline that assigns the value in trial $t$ to the average expert's future payoff as observed in the training set; and
\textbf{(e)} \textbf{History Proportion (HP)} -- a deterministic baseline which predicts that the future hotel choice rate is identical to the choice rate in previous steps.\footnote{In this baseline, as well as in the PD decision maker baseline, the past experiences are based on the gold standard.}


\textbf{\textit{AE.}} We compare our AE to ten alternatives, divided to four groups: (a-d) static rules; (e-g) dynamic rules, which adjust their predictions according to the behavior of the DM; (h) a greedy baseline which tests the VM classifier without the MCTS; and (i-j) variants of our original AE.

    \textbf{(a)} \textbf{RAND} -- an expert that randomly chooses a review from the available set;  \textbf{(b)} \textbf{MEDIAN} -- an expert that chooses the median scoring review at each trial.
    This baseline honestly communicates the value of the hotel;
   \textbf{(c)} \textbf{HIGHEST} -- an expert that chooses the highest scoring review at each trial. This expert always overestimates the value of the hotel; \textbf{(d)} \textbf{EXTREMIST} -- an expert that chooses the highest scoring review if the average review score is at least 8, and otherwise chooses the lowest scoring review. 
   This expert makes the strongest positive recommendation when the hotel crosses the "likely gain" threshold, and the strongest negative recommendation otherwise. 
   \textbf{(e)} \textbf{ADAPTIVE LIAR (A-LIAR)} -- An expert that reveals the highest scoring review as long as the DM keeps accepting the hotels. After the first rejection by the DM, the expert chooses randomly between the second and third highest scoring reviews. After the second rejection it reveals the median review for the remaining hotels; \textbf{(f+g)} \textbf{PERSONAL TASTE DETECTION (PTD)} -- this expert selects the review which is most similar to the average review representation, among the  hotels accepted in previous trials. We consider either the HC features (\textbf{PTD-HC}) or the BERT features (\textbf{PTD-BERT}) of the reviews, and compute similarity with the cosine operator;\footnote{In the first round the review is randomly selected.} \textbf{(h)} \textbf{VM SOFTMAX (VM-SM)} -- a greedy expert that at each trial selects a review with a probability proportional to the expected expert payoff associated with it according to the VM. This expert helps us quantify the added value of MCTS over a greedy strategy; \textbf{(i+j)} our AE when using the second best DMM (\textbf{AE-DM2}) and the second best VM (\textbf{AE-VM2}).



\paragraph{Numerical Communication}

The success of our AE depends both on our modeling approach and on the use of text-based communication between the expert and the DM. In order to separate the impact of these two characteristics, we replicate our experiments where the communication between the expert and the DM is purely numerical. To achieve this goal we utilize another dataset collected by \citet{apel2020predicting}. The authors collected data from 493 games (392 train and 101 test) with the same hotels and reviews discussed in $\S$ \ref{data} (including the split to training and test hotels), but with a different set of participants. In these numerical communication experiments the experts are presented with all seven reviews but are told that they can only reveal to the DM the score of one of them, rather than its text. The DM, in turn, decides whether or not to accept the hotel based solely on the revealed numerical score. Other than that the experimental setup in this condition is identical to that of the textual communication experiments.

This data allows us to test a numerical communication version of our AE. To this end we trained the following models: (a) \textit{DMM}: \textbf{SG-LSTM}: Our original LSTM-based DMM trained on the numerical communication training set, employing only the SG features; and (b) \textit{VM}: \textbf{SG-LSTM}: Our original LSTM-based VM trained on the numerical communication training set, employing only the SG features. Finally, we test the \textbf{AE-SG} model, an MCTS-based expert identical to our AE, except that it uses the SG-LSTM variants of the DMM and VM. The test setup is identical to the above, except that the simulations are based on the numerical communication DMM and VM.

\paragraph{Training Procedure and Hyper-parameters}

We apply a 5-fold cross validation protocol on the training set, and determine the optimal configuration of hyper-parameters according to the best average F1 score of the minority class -- hotel rejection. Next, we train the DMM and VM with their optimal configurations on the entire training set, and report results on the test set. 

For the HC-LSTM models we optimize the hidden layer size (64, 128, 256), the batch size (5, 10, 15, 20, 25) and the dropout value (0.3, 0.4, 0.5, 0.6). Training is carried out for 100 epochs with an early stopping criterion.
For the BERT-LSTM models we use HuggingFace's implementation of the pre-trained uncased BERT-Base model. \footnote{\url{https://github.com/huggingface/transformers}.} We tune the hidden layer size (64, 128, 256) and the dropout value (0.3,0.4,0.5,0.6) of the LSTM component, and set the batch size to 5. During the training of BERT-LSTM we keep BERT's parameters fixed for the first 8 epochs, and fine-tune them for additional 4 to 12 epochs with early stopping.

For MCTS we set the exploration constant \(c\) to 0.5, after normalizing the rewards to be in the [0,1] range, and the time limit constant to 1.5 minutes. Our AE uses the MCTS with the HC-LSTM variant for DMM and VM, which were selected in cross-validation experiments on the training data. Likewise, VM-SM uses the HC-LSTM model.

\com{
\paragraph{Evaluation Measures}
We now describe the evaluation measures for the DMM and VM:

\textbf{(a)} Predicting the DM's decision in each trial of the game is a binary classification task. Hence, we use an accuracy measure, for detecting the proportion of trials where the true decisions are predicted. In addition, we report macro-average F1 scores by averaging the F1 scores of each label.

\textbf{(b)} The VMs and their baseline predict the expert's future payoff in each trial of the game. Hence, we use the accuracy measure for detecting the proportion of trials correctly predicted. In addition, we report the root-mean-square-error (RMSE) to detect the absolute differences between the real and the predicted future payoffs.}

\section{Results}

This section present our results. We would first like (\S \ref{sec:sub-models}) to evaluate the performance of our DMM and VM models, since they are key elements of our AE. After verifying their quality, we turn to present our main results (\S \ref{sec:main-results}), comparing our AE to the various baselines. This will allow us to answer our three research questions (\S \ref{sec:introduction}), related to the AE performance (Q1), its adaptation to different decision maker types (Q2) and its strategy compared to humans (Q3).

\subsection{The DMM and VM Models}
\label{sec:sub-models}

\paragraph{DMM Results}

Table \ref{table:dmm_vm} (top) presents the accuracy and macro average F1-score results of the DMM variants on the binary task of predicting whether or not a human DM will choose to accept a suggested hotel. The results show that the best performing model is the HC-LSTM which yields an accuracy of 82.40\% and a macro average F1-score of 73.20. This result reflects the value of the hand-crafted textual features, a pattern that was also reported by \citet{apel2020predicting}. BERT-LSTM lags a bit behind (accuracy of 80.80\%, macro F1 score of 68.30), demonstrating that clever feature design can outperform this strong language encoder. 
In general, the SVM baselines fall short of the neural networks, whereas the deterministic baselines PD and EWG are not very successful.

\com{
\begin{table}[t!]
\centering
\resizebox{0.43\textwidth}{1.52cm}{
\begin{tabular}{ |p{2.4cm}|p{2.0cm}|p{2.0cm}|}
\hline
\textbf{Model}&\textbf{Accuracy $\uparrow$} & \textbf{F1-score $\uparrow$} \\
\hline
HC-LSTM & \textbf{82.40\%} & \textbf{73.20\%} \\
\hline
BERT-LSTM & 80.80\% & 68.30\% \\
\hline
HC-SVM & 79.50\%   & 68.50\% \\
\hline
BERT-SVM & 75.80\%   & 52.00\% \\
\hline
PD & 69.90\% & 45.21\%   \\
\hline
EWG & 60.00\% & 50.00\% \\
\hline
\end{tabular}}
\caption{Evaluation of DMM variants.}
\label{table:dm_results}
\end{table}

\begin{table}[t!]\label{value_results}
\centering
\resizebox{0.43\textwidth}{1.7cm}{

\begin{tabular}{ |p{2.4cm}|p{2.0cm}|p{2.0cm}|  }
\hline
\textbf{Model}&\textbf{Accuracy $\uparrow$} &\textbf{RMSE $\downarrow$}   \\
\hline

HC-LSTM &  \textbf{38.90\%} & 1.11 \\
\hline
BERT-LSTM & 16.70\% & 2.14 \\
\hline
HC-SVR & 35.40\% & 1.13 \\
\hline
BERT-SVR & 25.54\% & 1.41 \\
\hline
AVG &  33.70\% &\textbf{1.08}\\
\hline
DO & 26.20\% & 1.94\\
\hline
HP & 29.10\% & 1.90\\
\hline
\end{tabular}}
\caption{\label{value_results} Evaluation of VM variants.}
\end{table}}

\com{
\begin{table}[t!]
\centering
\resizebox{0.38\textwidth}{2.8cm}{
\begin{tabular}{ccc}
\hline
\multicolumn{1}{|c|}{\textbf{DMM}} & \multicolumn{1}{c|}{\textbf{Accuracy $\uparrow$}} & \multicolumn{1}{c|}{\textbf{F1-score $\uparrow$}} \\ \hline
\multicolumn{1}{|c|}{HC-LSTM}      & \multicolumn{1}{c|}{\textbf{82.40\%}}  & \multicolumn{1}{c|}{\textbf{73.20\%}}  \\ \hline
\multicolumn{1}{|c|}{BERT-LSTM}    & \multicolumn{1}{c|}{80.80\%}           & \multicolumn{1}{c|}{68.30\%}           \\ \hline
\multicolumn{1}{|c|}{HC-SVM}       & \multicolumn{1}{c|}{79.50\%}           & \multicolumn{1}{c|}{68.50\%}           \\ \hline
\multicolumn{1}{|c|}{BERT-SVM}     & \multicolumn{1}{c|}{75.80\%}           & \multicolumn{1}{c|}{52.00\%}           \\ \hline
\multicolumn{1}{|c|}{PD}           & \multicolumn{1}{c|}{69.90\%}           & \multicolumn{1}{c|}{45.21\%}           \\ \hline
\multicolumn{1}{|c|}{EWG}          & \multicolumn{1}{c|}{60.00\%}           & \multicolumn{1}{c|}{50.00\%}           \\ \hline
\multicolumn{1}{l}{}               & \multicolumn{1}{l}{}                   & \multicolumn{1}{l}{}                   \\ \hline
\multicolumn{1}{|c|}{\textbf{VM}}  & \multicolumn{1}{l|}{\textbf{Accuracy $\uparrow$}} & \multicolumn{1}{l|}{\textbf{RMSE $\downarrow$}}     \\ \hline
\multicolumn{1}{|c|}{HC-LSTM}      & \multicolumn{1}{c|}{\textbf{38.90\%}}  & \multicolumn{1}{c|}{1.11}              \\ \hline
\multicolumn{1}{|c|}{BERT-LSTM}    & \multicolumn{1}{c|}{16.70\%}           & \multicolumn{1}{c|}{2.14}              \\ \hline
\multicolumn{1}{|c|}{HC-SVR}       & \multicolumn{1}{c|}{35.40\%}           & \multicolumn{1}{c|}{1.13}              \\ \hline
\multicolumn{1}{|c|}{BERT-SVR}     & \multicolumn{1}{c|}{25.54\%}           & \multicolumn{1}{c|}{1.41}              \\ \hline
\multicolumn{1}{|c|}{AVG}          & \multicolumn{1}{c|}{33.70\%}           & \multicolumn{1}{c|}{\textbf{1.08}}     \\ \hline
\multicolumn{1}{|c|}{DO}           & \multicolumn{1}{c|}{26.20\%}           & \multicolumn{1}{c|}{1.94}              \\ \hline
\multicolumn{1}{|c|}{HP}           & \multicolumn{1}{c|}{29.10\%}           & \multicolumn{1}{c|}{1.90}              \\ \hline
\end{tabular}}
\caption{\label{table:dmm_vm} Evaluation of DMM and VM variants.}
\end{table}}

\begin{table}[t!]
\centering
\scalebox{0.65}{
\begin{tabular}{ccc}
\hline
\multicolumn{1}{|c|}{\textbf{DMM}} & \multicolumn{1}{c|}{\textbf{Accuracy $\uparrow$}} & \multicolumn{1}{c|}{\textbf{F1-score $\uparrow$}} \\ \hline
\multicolumn{1}{|c|}{HC-LSTM}      & \multicolumn{1}{c|}{\textbf{82.40\%}}  & \multicolumn{1}{c|}{\textbf{73.20}}  \\ \hline
\multicolumn{1}{|c|}{BERT-LSTM}    & \multicolumn{1}{c|}{80.80\%}           & \multicolumn{1}{c|}{68.30}           \\ \hline
\multicolumn{1}{|c|}{SG-LSTM}          & \multicolumn{1}{c|}{77.00\%}           & \multicolumn{1}{c|}{65.70}           \\ \hline
\multicolumn{1}{|c|}{HC-SVM}       & \multicolumn{1}{c|}{79.50\%}           & \multicolumn{1}{c|}{68.50}           \\ \hline
\multicolumn{1}{|c|}{BERT-SVM}     & \multicolumn{1}{c|}{75.80\%}           & \multicolumn{1}{c|}{52.00}           \\ \hline
\multicolumn{1}{|c|}{PD}           & \multicolumn{1}{c|}{69.90\%}           & \multicolumn{1}{c|}{45.21}           \\ \hline
\multicolumn{1}{|c|}{EWG}          & \multicolumn{1}{c|}{60.00\%}           & \multicolumn{1}{c|}{50.00}           \\ \hline
\multicolumn{1}{l}{}               & \multicolumn{1}{l}{}                   & \multicolumn{1}{l}{}                   \\ \hline
\multicolumn{1}{|c|}{\textbf{VM}}  & \multicolumn{1}{l|}{\textbf{Accuracy $\uparrow$}} & \multicolumn{1}{l|}{\textbf{RMSE $\downarrow$}}     \\ \hline
\multicolumn{1}{|c|}{HC-LSTM}      & \multicolumn{1}{c|}{\textbf{38.90\%}}  & \multicolumn{1}{c|}{1.11}              \\ \hline
\multicolumn{1}{|c|}{BERT-LSTM}    & \multicolumn{1}{c|}{16.70\%}           & \multicolumn{1}{c|}{2.14}              \\ \hline
\multicolumn{1}{|c|}{SG-LSTM}    & \multicolumn{1}{c|}{33.95\%}           & \multicolumn{1}{c|}{1.40}              \\ \hline
\multicolumn{1}{|c|}{HC-SVR}       & \multicolumn{1}{c|}{35.40\%}           & \multicolumn{1}{c|}{1.13}              \\ \hline
\multicolumn{1}{|c|}{BERT-SVR}     & \multicolumn{1}{c|}{25.54\%}           & \multicolumn{1}{c|}{1.41}              \\ \hline
\multicolumn{1}{|c|}{AVG}          & \multicolumn{1}{c|}{33.70\%}           & \multicolumn{1}{c|}{\textbf{1.08}}     \\ \hline
\multicolumn{1}{|c|}{DO}           & \multicolumn{1}{c|}{26.20\%}           & \multicolumn{1}{c|}{1.94}              \\ \hline
\multicolumn{1}{|c|}{HP}           & \multicolumn{1}{c|}{29.10\%}           & \multicolumn{1}{c|}{1.90}              \\ \hline
\end{tabular}}
\caption{\label{table:dmm_vm} Evaluation of DMM and VM variants.}
\end{table}

\com{
\begin{table*}[!h]\label{AE_results}
\centering
\resizebox{0.88\textwidth}{2.45cm}{
\begin{tabular*}{\textwidth}{
|p{2cm}|p{1.4cm}|p{1.4cm}|p{1.7cm}|p{1.7cm}|p{1.7cm}|p{1.7cm}|p{0.92cm}|}
\hline
\textbf{Expert\textbackslash  DM} & \textbf{HC-LSTM} & \textbf{BERT-LSTM} &\textbf{HC-LSTM+0.1}&\textbf{HC-LSTM+0.2}&\textbf{HC-LSTM-0.1}&\textbf{HC-LSTM-0.2}&\textbf{AVG}\\
\hline
\hline


\hline
AE & \textbf{7.12}& 7.04 & \textbf{8.10} & 8.77& \textbf{6.04} & \textbf{5.02} & \textbf{7.02} \\
\hline
\hline
RAND & 6.54 & 6.67 & 7.56 & 8.31 & 5.58 & 4.49 & 6.53 \\
\hline
MEDIAN & 6.46 & 6.85 & 7.24 & 8.02 & 5.45 & 4.66 & 6.45 \\
\hline
HIGHEST &  6.77  &  \textbf{7.82} & 7.94 & \textbf{8.82} & 5.55 & 4.46 & 6.89 \\
\hline
EXTREMIST & 6.21 &  6.86 & 7.24 & 7.99 & 5.14 & 4.08 & 6.25 \\
\hline
\hline
A-LIAR & 6.54 & 7.14 & 7.15 & 8.69 & 5.40 & 4.35 & 6.55 \\
\hline
PTD-HC & 6.88 & 7.03 & 7.68 & 8.49 & 5.83 & 4.92 & 6.83 \\
\hline
PTD-BERT & 6.79 & 6.59 & 7.72 & 8.46 & 5.77 & 4.82 & 6.69 \\
\hline
\hline
VM-SM & 6.58 & 7.00 & 7.70 & 8.34 & 5.65 & 4.67 & 6.66 \\
\hline
AE-DM2 & 7.05 & 7.23 & 7.94 & 8.66 & 5.92 & 4.97 & 6.96 \\
\hline
AE-VM2 & 7.03 & 7.05 & 8.00 & 8.76 & 5.98 & 4.98 & 6.97 \\
 \hline
AE-SG & 6.98 & 7.07 & 7.91 & 8.64 & 5.92 & 4.91 & 6.91 \\
\hline
\end{tabular*}}
\caption{Average expert's payoff over 1000 simulations against different DMs.
The table is split into four sections, from top to bottom: Our model (AE), static rules, dynamic rules and algorithms.
The human experts in the experiments of \citet{apel2020predicting} achieve an average payoff of 7.36.}
\label{table:expert_results}
\end{table*}}

\com{
\begin{table*}[!h]\label{AE_results}
\centering
\scalebox{0.7}{
\begin{tabular*}{\textwidth}{
|p{2cm}|p{1.4cm}|p{1.4cm}|p{1.7cm}|p{1.7cm}|p{1.7cm}|p{1.7cm}|p{0.92cm}|}
\hline
\textbf{Expert\textbackslash  DM} & \textbf{HC-LSTM} & \textbf{BERT-LSTM} &\textbf{HC-LSTM+0.1}&\textbf{HC-LSTM+0.2}&\textbf{HC-LSTM-0.1}&\textbf{HC-LSTM-0.2}&\textbf{AVG}\\
\hline
\hline


\hline
AE & \textbf{7.12}& 7.04 & \textbf{8.10} & 8.77& \textbf{6.04} & \textbf{5.02} & \textbf{7.02} \\
 & [7.02, 7.22] & [7.03, 7.29] & [8.02, 8.19] & [8.70, 8.84] & [5.93, 6.20] & [4.90, 5.13] &  \\
\hline
\hline
RAND & 6.54 & 6.67 & 7.56 & 8.31 & 5.58 & 4.49 & 6.53 \\
 & [6.49, 6.7] & [6.56, 6.77]& [7.47, 7.65]&
[8.24, 8.38] & [5.47, 5.68]&
[4.38, 4.60]&  \\
\hline
MEDIAN & 6.46 & 6.85 & 7.24 & 8.02 & 5.45 & 4.66 & 6.45 \\
& [6.37, 6.54]& [6.76, 6.96] & [7.16, 7.33] & [7.96, 8.11] & [5.37, 5.54] & [4.56, 4.76] &  \\
\hline
HIGHEST &  6.77  &  \textbf{7.82} & 7.94 & \textbf{8.82} & 5.55 & 4.46 & 6.89 \\
 & [6.65, 6.89] & [7.73, 7.92] & [7.84, 8.04] & [8.74, 8.89] & [5.42, 5.68]& [4.33, 4.58] & \\ 
\hline
EXTREMIST & 6.21 &  6.86 & 7.24 & 7.99 & 5.14 & 4.08 & 6.25 \\
& [6.11, 6.32] & [6.76, 6.96] & [7.14,7.34] & [7.92, 8.09] & [5.04, 5.26] & [3.97, 4.19] & \\
\hline
\hline
A-LIAR & 6.54 & 7.14 & 7.15 & 8.69 & 5.40 & 4.35 & 6.55 \\
& [6.42, 6.65] & [7.06, 7.28] & [7.06, 7.28]& [8.61, 8.77] & [5.28, 5.51] & [4.24, 4.47] & \\
\hline
PTD-HC & 6.88 & 7.03 & 7.68 & 8.49 & 5.83 & 4.92 & 6.83 \\
& [6.78, 6.99] &[6.86, 7.06] &[7.63, 7.80] &[8.43, 8.57] &[5.72, 5.95] &[4.79, 5.02]& \\
\hline
PTD-BERT & 6.79 & 6.59 & 7.72 & 8.46 & 5.77 & 4.82 & 6.69 \\
& [6.67, 6.88] &[6.51, 6.73] &[7.63, 7.82] &[8.38, 8.54] &[5.64, 5.88] & [4.71, 4.93] & \\
\hline
\hline
VM-SM & 6.58 & 7.00 & 7.70 & 8.34 & 5.65 & 4.67 & 6.66 \\
& [6.50, 6.71] & [6.91, 7.12] &[7.60, 7.77] & [8.26, 8.41] & [5.58, 5.79] &[4.57, 4.80] & \\
\hline
AE-DM2 & 7.05 & 7.23 & 7.94 & 8.66 & 5.92 & 4.97 & 6.96 \\
& [6.93, 7.14] & [7.13, 7.33] & [7.86, 8.02] & [8.58, 8.73] & [5.84, 6.07] &[4.89, 5.10] & \\
\hline
AE-VM2 & 7.03 & 7.05 & 8.00 & 8.76 & 5.98 & 4.98 & 6.97 \\
& [6.93, 7.13] & [6.96, 7.17] & [7.93, 8.09] & [8.72, 8.85] & [5.88, 6.09] & [4.90, 5.12] & \\
\hline
\hline
\hline
AE-SG & 7.53 & - & 8.63 & 9.10 & 6.02 & 4.85 & 7.23 \\
& [7.39, 7.64]&
- & [8.48, 8.65] & [9.09, 9.22] & [5.93, 6.23] & [4.61, 4.93] & \\
\hline
\end{tabular*}}
\caption{Average expert's payoff over 1000 simulations against different DMs.
The table is split into five sections, from top to bottom: Our model (AE), static rules, dynamic rules, algorithms, and the results in the numerical communication setup, which are not directly comparable to the above, text-based communication results. For each condition, we report the average expert payoff over our 1000 simulations, as well as 95\% CI (in brackets, using bootstrap re-sampling with 1000 re-samples of our original 1000 simulations; see \citet{dror2018hitchhiker}).
The human experts in the experiments of \citet{apel2020predicting} achieve an average payoff of 7.36.}
\label{table:expert_results}
\end{table*}}

\begin{table*}[!h]\label{AE_results}
\centering
\scalebox{0.73}{
\begin{tabular}{|c|c|c|c|c|c|c|c|}
\hline
Expert/DM & HC-LSTM & BERT-LSTM & HC-LSTM+0.1 & HC-LSTM+0.2 & HC-LSTM-0.1 & HC-LSTM-0.2 & AVG  \\ \hline
AE        & \textbf{7.12} [7.02, 7.22]               & 7.04 [7.03, 7.29]                 & \textbf{8.10} [8.02, 8.19]                   & 8.77 [8.70, 8.84]                   & \textbf{6.04} [5.93, 6.20]                   & \textbf{5.02} [4.90, 5.13]                   & \textbf{7.02} \\ \hline \hline
RAND      & 6.54 [6.49, 6.70]                & 6.67 [6.56, 6.77]                 & 7.56 [7.47, 7.65]                   & 8.31 [8.24, 8.38]                   & 5.58 [5.47, 5.68]                   & 4.49 [4.38, 4.60]                   & 6.53 \\ \hline
MEDIAN    & 6.46 [6.37, 6.54]               & 6.85 [6.76, 6.96]                 & 7.24  [7.16, 7.33]                  & 8.02 [7.96, 8.11]                   & 5.45 [5.37, 5.54]                   & 4.66 [4.56, 4.76]                   & 6.45 \\ \hline
HIGHEST   & 6.77 [6.65, 6.89]               & \textbf{7.82} [7.73, 7.92]                 & 7.94 [7.84, 8.04]                   & \textbf{8.82} [8.74, 8.89]                   & 5.55 [5.42, 5.68]                   & 4.46 [4.33, 4.58]                   & 6.89 \\ \hline
EXTREMIST & 6.21 [6.11, 6.32]               & 6.86 [6.76, 6.96]                 & 7.24 [7.14,7.34]                    & 7.99 [7.92, 8.09]                   & 5.14 [5.04, 5.26]                   & 4.08 [3.97, 4.19]                   & 6.25 \\ \hline \hline
A-LIAR    & 6.54 [6.42, 6.65]               & 7.14 [7.06, 7.28]                 & 7.15 [7.06, 7.28]                   & 8.69 [8.61, 8.77]                   & 5.40 [5.28, 5.51]                   & 4.35 [4.24, 4.47]                   & 6.55 \\ \hline
PTD-HC    & 6.88 [6.78, 6.99]               & 7.03 [6.86, 7.06]                 & 7.68 [7.63, 7.80]                   & 8.49 [8.43, 8.57]                   & 5.83 [5.72, 5.95]                   & 4.92 [4.79, 5.02]                   & 6.83 \\ \hline
PTD-BERT  & 6.79 [6.67, 6.88]               & 6.59 [6.51, 6.73]                 & 7.72 [7.63, 7.82]                   & 8.46 [8.38, 8.54]                   & 5.77 [5.64, 5.88]                   & 4.82 [4.71, 4.93]                   & 6.69 \\ \hline \hline
VM-SM     & 6.58 [6.50, 6.71]               & 7.00 [6.91, 7.12]                 & 7.70 [7.60, 7.77]                   & 8.34 [8.26, 8.41]                   & 5.65 [5.58, 5.79]                   & 4.67 [4.57, 4.80]                   & 6.66 \\ \hline
AE-DM2    & 7.05 [6.93, 7.14]               & 7.23 [7.13, 7.33]                 & 7.94 [7.86, 8.02]                   & 8.66 [8.58, 8.73]                   & 5.92 [5.84, 6.07]                   & 4.97 [4.89, 5.10]                   & 6.96 \\ \hline
AE-VM2    & 7.03 [6.93, 7.13]               & 7.05 [6.96, 7.17]                 & 8.00 [7.93, 8.09]                   & 8.76 [8.72, 8.85]                   & 5.98 [5.88, 6.09]                   & 4.98 [4.90, 5.12]                   & 6.97 \\ \hline \hline
AE-SG     & 7.53 [7.39, 7.64]               & -                                 & 8.63 [8.48, 8.65]                   & 9.10 [9.09, 9.22]                   & 6.02 [5.93, 6.23] & 4.85 [4.61, 4.93]               & 7.23    \\ \hline
\end{tabular}}
\caption{Average expert's payoff over 1000 simulations against different DMs.
The table is split into five sections, from top to bottom: Our model (AE), static rules, dynamic rules, algorithms, and the results in the numerical communication setup, which are not directly comparable to the above, text-based communication results. For each condition, we report the average expert payoff over our 1000 simulations, as well as 95\% CI (in brackets, using bootstrap re-sampling with 1000 re-samples of our original 1000 simulations; see \citet{dror2018hitchhiker}).
The human experts in the experiments of \citet{apel2020predicting} achieve an average payoff of 7.36.}
\label{table:expert_results}
\end{table*}

\com{
\begin{table}[t!]
\begin{tabular}{|l|l|}
\hline
Expert    & Average DM Payoff \\ \hline
AE        & 2.37     \\ \hline
RAND      & 2.60     \\ \hline
MEDIAN    & \textbf{3.50}       \\ \hline
HIGHEST   & 2.13             \\ \hline
EXTREMIST & 2.94              \\ \hline
A-LIAR    & 2.63       \\ \hline
PTD-HC    & 2.67       \\ \hline
PTD-BERT  & 2.53       \\ \hline
VM-SM     & 2.67       \\ \hline
AE-VM2    & 2.36       \\ \hline
AE-DM2    & 2.26              \\ \hline
AE-SG    & 2.35              \\ \hline
\end{tabular}
\caption{Average DM payoff achieved against each of the experts.}
\label{table:dm_avg_payoff}
\end{table}
}

\paragraph{VM Results}

Table \ref{table:dmm_vm} (bottom) presents the exact accuracy and Root Mean Square Error (RMSE) of the VM variants on the task of predicting the experts' future payoff. The strongest model is HC-LSTM (best exact accuracy, second best RMSE). Moreover, the second best model is HC-SVR, which also exploits the hand-crafted textual features. In contrast, the BERT-based models perform quite poorly. This illustrates once again the strong positive impact of the HC features, that are very effective even when the task classifier does not model the structure of the data. Interestingly, the same features and architecture perform best both for the DMM and for the VM.

The AVG baseline, which always predicts the average score, obtains the lowest RMSE score, but it is not as accurate as our HC-based models. DO and HP, that are based on simple statistical rules, also perform quite poorly.


\subsection{Main Results: Automated Expert Performance against Different DMs}
\label{sec:main-results}

Table \ref{table:expert_results} presents AE results (averaged over 1000 simulated games) when playing with 6 different DMs. 
Notice that our AE employs the HC-LSTM based DMM and VM variants at all times -- the columns of the table correspond to the different DMs it plays with. Recall that the AE can adapt itself to its rival through the statistical game features, which reflect the behavior of the rival DM at previous trials. This allows us to test how well our AE generalizes to new players with different strategies than those it assumes.

The results suggest that our AE is the best expert, reaching the best average payoff overall, the best average payoff when playing against 4 of the 6 DMs, and the second and fifth best payoffs when playing against the remaining 2 DMs. These encouraging results indicate the capability of our AE to adapt itself to various DM types, providing a positive answer for Q1 and Q2. 

The human experts  in the experiments of \citet{apel2020predicting} achieved an average payoff of 7.36, somewhat higher than the 7.02 average of our AE. Note, however, that the human experts of \citet{apel2020predicting} played against human DMs and hence the results are not directly comparable. Yet, hoping that the various automated DMs provide a representation of the prominent types of human DMs, we consider the small gap between the two numbers to provide an optimistic indication that the answer to Q3 may be positive and our AE performs similarly to human experts, at least with respect to its payoff. Below (\S \ref{sec:ablation}) we further analyse the choices made by our AE, demonstrating interesting properties of its revealed texts and comparing its decisions to those of the human experts of \citet{apel2020predicting}.

Interestingly, the HIGHEST baseline performs best and third-best, respectively, against HC-LSTM+0.2 and HC-LSTM+0.1. This is because these compromised DMs tend to accept the hotel for almost every review that they are presented with. However, for HC-LSTM, and for the inflexible DMs, HC-LSTM-0.1 and HC-LSTM-0.2, HIGHEST is far from being the best model.


Additionally, the EXTREMIST and MEDIAN baselines, which aim to select the review that best reflects the different hotel scores, are inferior to our AE in all setups. Two possible explanations can be considered. Firstly, unlike the AE that is trained to maximize its payoff, EXTREMIST and MEDIAN favor the DM by being transparent in their choices at the expense of their own benefits. Secondly, unlike the AE, these baselines do not exploit the textual features of the reviews. The strong performance of the AE is an indication of the importance of textual features for strategy design.

Finally, the dynamic rules (A-LIAR, PTD-HC and PTD-BERT), the greedy VM-SM, and the AE-DM2 and AE-VM2 versions of our AE, which use the second best DMM (BERT-LSTM) or VM (HC-SVR), respectively, are inferior to our AE. We consider this an indication of the importance of a wise search procedure, that carefully balances the long (explore) and the short (exploit) terms, and of careful selection of suitable DMM and VM.

\begin{figure}[!t]
  \centering
    \includegraphics[width=0.5\textwidth]{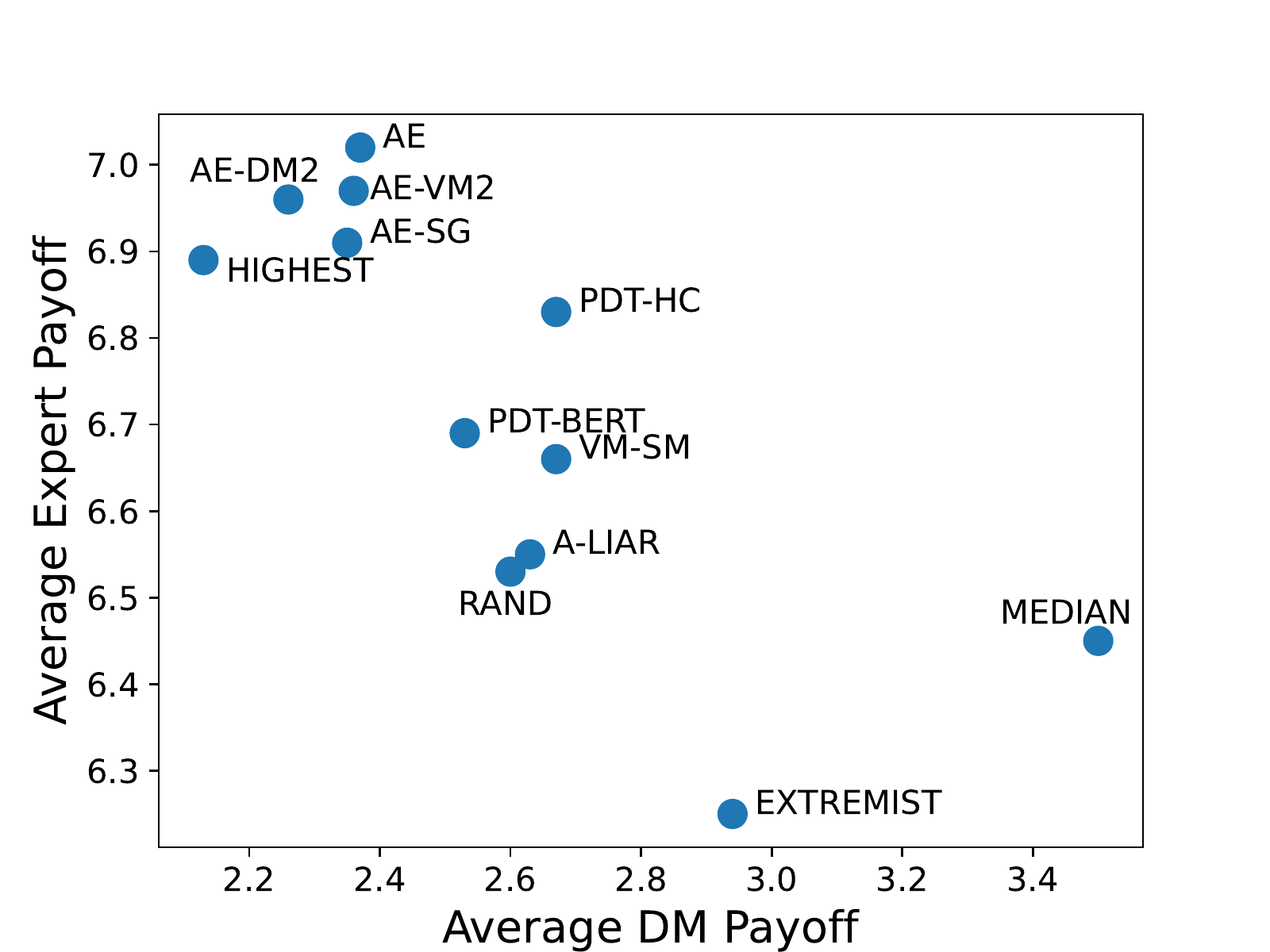}
  \caption{Average expert payoff as a function of the average payoff of the DMs it played with.}
\label{fig:expert_vs_dm_avg_payoff}
\end{figure}

\section{Ablation Analysis}
\label{sec:ablation}

In this section we analyse several aspects of the main results presented above. We start by analysing the impact of text-based communication on our results, evaluating the performance of our AE when performing numerical communication. Then, we analyse different aspects of the observed behavior of our AE in our main text-based communication experiments: The average payoff of the DMs (indicating whether our AE facilitates fairness), the decision patterns of our AE when playing against the various DMs (shedding more light on Q2 - does the AE adapt to the DM it plays with), comparing the reviews revealed by our AE to those revealed by the human experts in \citet{apel2020predicting} (thus shedding more light on Q3), and, finally, analysing the textual properties of the reviews revealed by our AE. 

\paragraph{Numerical Communication Results}

To put our textual communication results in context, we also report results for the numerical communication setup (Table \ref{table:expert_results}, bottom line). As above, we report results for the DMM and VM models, based on the SG-LSTM architecture, and for the eventual AE-SG expert. We cannot compare these results directly to the textual communication numbers, as they are based on another set of games and a different type of communication, but we do hope to learn about the difference between the communication types based on the observed patterns.

The numerical communication DMM:SG-LSTM and VM:SG-LSTM models achieve accuracy scores of 77.00\% and 33.95\%, respectively. The F1-score of the DMM is 65.70 and the RMSE score of the VM is 1.4. Interestingly, these numbers are substantially lower than the comparable numbers of the leading textual communication models (see Table~\ref{table:dmm_vm}). This is an indication that it is harder to predict the DM behavior as well as the future AE payoff when the communication is numerical and hence only behavioral features can be used for prediction. 

Interestingly, the AE-SG model achieved payoffs of 7.53, 8.63, 9.1, 6.02 and 4.85 against the numerical HC-LSTM, HC-LSTM+0.1, HC-LSTM+0.2, HC-LSTM-0.1 and HC-LSTM-0.2, respectively (there is no BERT-LSTM simulation when communication is numerical). These payoffs are higher than the best AE payoff in the textual communication setups in the first 3 cases, but are lower in the last 2 setups where the acceptance probability of the simulated DM is decreased. 

While this comparison between numerical and textual communication is interesting, we notice that in many real-life scenarios the communication is either numerical or verbal. Hence, it is important to design effective models for both cases. 

\paragraph{Average DM Payoff}

Figure \ref{fig:expert_vs_dm_avg_payoff} presents the average payoff of each expert as a function of the average payoff of the DMs it played with. The figure suggests that DMs who played with the two experts with the lowest average payoff (MEDIAN and EXTREMIST) achieve the highest payoff on average. Our AE, in contrast, the highest-paid expert on average, leads to one of the lowest average DM payoffs. Generally, we observe a strong negative correlation of -0.76 between the average payoffs of the expert and the DM. As discussed in \S \ref{sec:introduction}, our game is not a zero-sum game; Yet, the negative correlation between the payoffs of the expert and the DM, even for experts that were not trained to maximize their own payoffs (like our AE and the numerical communication AE-SG), demonstrates the competitive nature of our task. A major goal of future research is to design an expert that can balance the payoffs of the two players, ideally maximizing them at the same time.

\paragraph{Analysis of AE Personalization}

\com{One of the most desirable characteristics of the AE is the ability to personalize its decisions based on the DM it faces against. We decide to capture such behaviour by measuring the average review score that our AE chooses to reveal during the 1000 simulations for the five DMs presented in the experiments of Table \ref{table:expert_results}. The different DM variants represent a wide range of compromise tendencies: inflexible (HC-LSTM-0.1, HC-LSTM-0.2), neutral (HC-LSTM) and compromised (HC-LSTM+0.1, HC-LSTM+0.2). Figure \ref{fig:compromise_tendency} presents the average review score selected for each of the DM, after normalizing the scores according to the min-max normalization. The figure suggests that there is a strong positive correlation between the scores our AE decided to reveal to the tendency to compromise. The higher the tendency of the DM to accept hotels, the higher are its revealed scores. This favorable behaviour of our AE serves as an evidence to its generalizability and to the confirmation of Question \ref{question:generalizability}.}
\com{\begin{figure}[!h]
\includegraphics[width=8.6cm]{figures/avg_revealed_scores.png}
  
\caption{The average score attached to the revealed reviews in the simulated game between the AE and each of the HC-LSTM DMM variants. The scores are normalized to the 0-1 range using the min-max normalization.}
        \label{fig:compromise_tendency} 
\end{figure}}

A desirable characteristic of an AE is the ability to personalize its decisions to the DM it faces. We analyse this behaviour by measuring the average review score that our AE chooses to reveal to the five HC-LSTM variants of Table \ref{table:expert_results}. 

Our analysis reveals that the higher the tendency of the DM to accept hotels, the higher are the scores of the reviews sent by the AE. We normalize the scores of each hotel to the $[0,1]$ range and compute the average review score selected across all hotels, for each of the DMs. The average scores are 0.483 (HC-LSTM-0.2), 0.485 (HC-LSTM-0.1), 0.487 (HC-LSTM), 0.488 (HC-LSTM+0.1) and 0.491 (HC-LSTM+0.2). This favorable behaviour of our AE serves as an evidence to its generalizability (Q2).

\com{
\begin{figure*}[!h]
\resizebox{\textwidth}{3.5cm}{
  \centering
    \includegraphics[width=\textwidth]{figures/ae_vs_he_10.png}}
  \caption{Revealed review score distribution for the AE and the human experts (HEs). The reviews are grouped into three bins according to their attached score: low (L), medium (M) or high (H), and the average score of the reviews in each group is in parentheses.}
        \label{fig:ae_vs_humans}
\end{figure*}}

\begin{figure}[!h]
  \centering
    \includegraphics[height=6.5cm,width=0.5\textwidth]{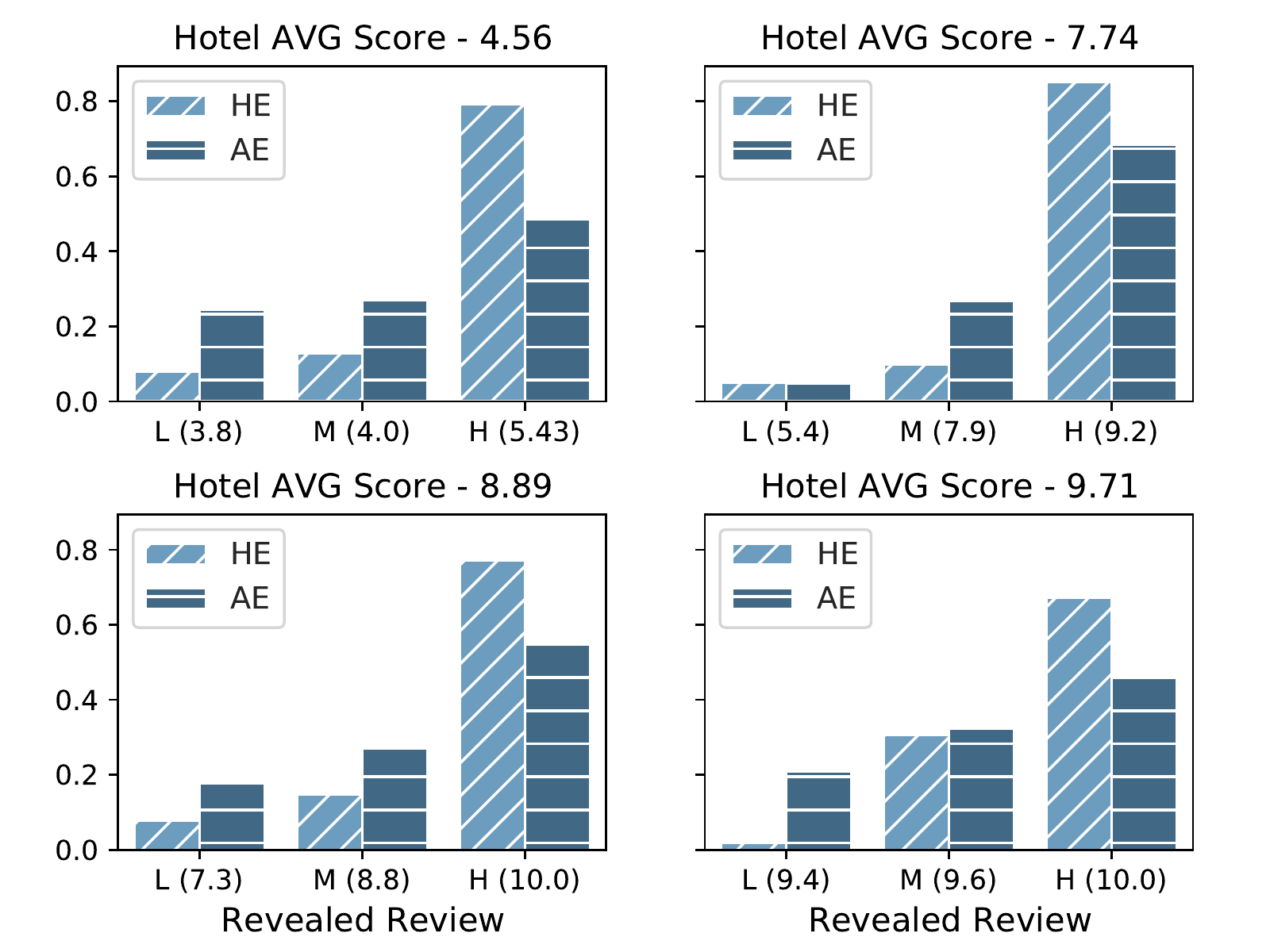}
  \caption{Revealed review score distributions for the AE and the human experts (HEs), for four representative hotels. The reviews are grouped into three bins according to their attached score: Low (L), medium (M) or high (H), and the average score of the reviews in each bin is in parentheses.}
        \label{fig:ae_vs_humans}
\end{figure}

\paragraph{AE vs. Human Experts}
One of the most interesting aspects of designing an AE is its similarity to human experts (HEs). To address this aspect (Q3), we compare between the AE and the HEs that participated in the experiments of \citet{apel2020predicting}.
Notice that the HEs play against human DMs, while our AE plays against artificial DMs, which makes them not directly comparable.

Figure \ref{fig:ae_vs_humans} depicts the score distributions of the reviews as revealed by the AE and the HEs for 4 representative test set hotels. We cluster the scores per hotel into 3 bins: Low, medium and high, and present the average score of each bin. The figure indicates that both experts consistently prefer to present highly ranked reviews and tend to reveal reviews that overestimate the hotels' average scores. Nonetheless, in all cases, the HEs output higher estimations, whereas the AE's scores are more diverse and closer to the average review score.
This analysis sheds light on our AE's behaviour, providing an initial answer to Q3. 

\begin{table}[!h]\label{hcfeatures}
  \centering
\scalebox{0.55}{
\begin{tabular}{ |P{3.9cm}|P{3.9cm}|P{3.9cm}|  }

\hline
\textbf{Low Scoring Hotels} &\textbf{Medium Scoring Hotels}&\textbf{High Scoring Hotels}   \\
\hline

 Location-Positive (92.5\%) &Room-Positive \hspace{1cm} (67.6\%) & Staff-Positive \hspace{1cm}(81.3\%)\\
\hline
Metro-Positive \hspace{1cm}(52.5\%)&Staff-Positive \hspace{1cm}(64.2\%)& Location-Positive\hspace{1cm} (74.9\%)\\
\hline
Staff-Positive\hspace{1cm} (46.8\%)&  Location-Positive (48.9\%) & Room-Positive\hspace{1cm} (47.3\%) \\
\hline
Staff-Negative \hspace{1cm}(45.0\%) & Metro-Positive \hspace{1cm}(38.2\%) & Facilities-Positive (29.4\%)\\
\hline
Facilities-Negative (44.6\%) &  Facilities-Negative (31.2\%)& Metro-Positive\hspace{1cm} (23.9\%)\\
\hline
\end{tabular}}
\caption{The top 5 topics (ordered by frequency) discussed in the reviews revealed by the AE for low, medium or high scoring hotels.}
\label{topics_results}
\end{table}

\paragraph{Textual Analysis of the AE-revealed Reviews}
We also analyze the textual features of the reviews that our AE chose when played against the LSTM-HC DM. Table \ref{topics_results} presents the top 5 topics discussed in the revealed reviews for low (average score $(as) < 7.5$), medium ($7.5 \leq as \leq 8.5$) or high ($as > 8.5$) scoring hotels. The topics are based on the HC features, that encode topics such as \textit{facilities, staff, location, food, design}, and \textit{price}, which are reviewed positively or negatively.

Interestingly, \textit{location}, \textit{staff}, and \textit{metro} are all discussed positively in the revealed reviews of the three hotel groups. However, the lower the hotel score is, the lower the rank of its \textit{staff} and the higher the rank of the \textit{metro}, among the top 5 topics. It hence seems that for low-scoring hotels the AE communicates positive aspects of their outer surroundings. 
%
Negative topics are more discussed in low and medium scoring hotels, with \textit{facilities} being negatively discussed in many revealed reviews of low-scoring and medium-scoring hotels.



\section{Human Experiments}

\begin{figure*}[!ht]
    \centering
    \includegraphics[width=\textwidth]{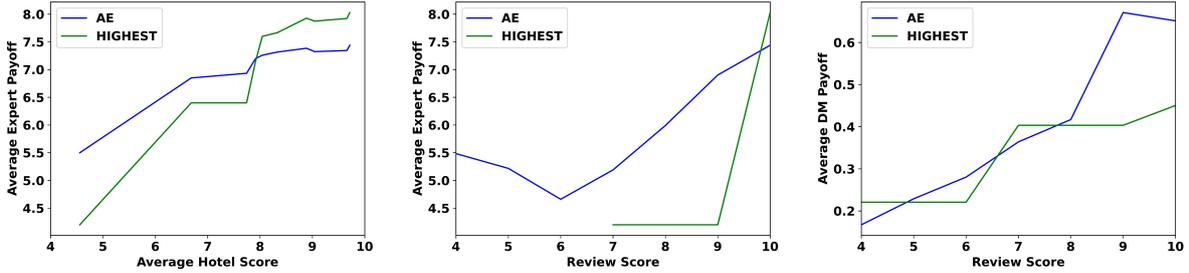}
    \caption{Average AE payoff for average hotel scores (\textbf{Left}) and for revealed review scores (\textbf{Middle}) that are \textbf{up to} a certain threshold (X-axis). (\textbf{Right}) Average DM payoff for \textbf{revealed} review scores that are \textbf{at least} of a certain threshold (X-axis).} 
    \label{fig:יhuman_experiments_grid}
\end{figure*}

Finally, we evaluate our AE when playing with human DMs. We do believe that simulation-based evaluation is crucial for our setting as it allows us to test our AE against DMs with a variety of controlled characteristics at a relatively low-cost (see \S \ref{sec:experiments}). Yet, human-based evaluation, even if it is small-scale due to its high cost, provides important complementary information.

\begin{table}[!t]
\centering
\scalebox{0.6}{
\begin{tabular}{|l|l|l|l|}
\hline
\textbf{}        & \multicolumn{3}{c|}{\textbf{All Games}}                           \\ \hline
\textbf{}        & \textbf{Expert Payoff} & \textbf{DM Payoff} & \textbf{Num. Players} \\ \hline
\textbf{AE}      & 7.44                   & 1.03               & 100                   \\ \hline
\textbf{HIGHEST} & \textbf{8.03}                   & \textbf{2.21}               & 100                   \\ \hline
\textbf{}        & \multicolumn{3}{c|}{\textbf{Acceptance Rate $\leq$ 80\%}}      \\ \hline
\textbf{}        & \textbf{Expert Payoff} & \textbf{DM Payoff} & \textbf{Num. Players} \\ \hline
\textbf{AE}      & 6.51                   & 0.70               & 70                    \\ \hline
\textbf{HIGHEST} & \textbf{6.60}                   & \textbf{0.72}               & 50                    \\ \hline
\end{tabular}
}
\caption{Average payoffs for all games (top) and when the acceptance rate $\leq$ 80\% (bottom).}
\label{table:average_dm_vs_expert_payoffs}
\end{table}

Following \citet{apel2020predicting}, our AE plays with 100 different human DMs on the Amazon Mechanical Turk (AMT) platform,\footnote{\url{https://www.mturk.com}} such that no DM competes against more than one expert.\footnote{We followed the exact same AMT experimental setup as in \citet{apel2020predicting}. Particularly, we filtered the AMT workers according to the two attention checks described in Section 4.1 of their paper.} We follow the same experimental setting as in our simulations, and particularly use the same test-set hotels.
We compare the performance of our AE to those of the strongest alternative: HIGHEST, the second best baseline in our simulations (in terms of average performance; the various AE agents are not considered as baselines in this definition).

Figure \ref{fig:יhuman_experiments_grid} (\textbf{Left}) illustrates the average expert payoff for hotels with an average review score of at most $s \in \{4,\ldots, 10\}$. The results suggest that our AE achieves the highest average payoffs for the 4 hotels with the lowest average review score (average score of up to 8), i.e., the hotels for which the expected DM payoff is negative. This observation implies that our AE is able to maximize its payoff on the most challenging hotels. The HIGHEST agent excels on the other 6 hotels, those with an average review score higher than 8 and hence a positive expected DM payoff. Interestingly, 5 of these 6 hotels have a review with the maximal score of 10, which is chosen by HIGHEST. 


We next analyse the scores of the revealed reviews -- i.e., the reviews that were chosen by the experts and presented to the DMs. Figure \ref{fig:יhuman_experiments_grid} (\textbf{Middle}) presents the average expert payoff when its revealed review score is at most $s \in \{4,\ldots, 10\}$. While the HIGHEST agent achieves the best payoff when it reveals a review with the maximal score of 10, when moving to lower scores we see that our agent maintains a higher average payoff. For such cases where the hotel does not have any review with the score of 10, the HIGHEST agent achieves a low average payoff of 4.2. 


The final analysis (Figure \ref{fig:יhuman_experiments_grid} (\textbf{Right})) is similar to first two, except that now we are focusing on the average DM payoff, when the revealed review score is at least $s \in \{4,\ldots, 10\}$. The leftmost point, corresponding to all experiments, suggests that in total the human DMs who played with our AE achieve the lowest average payoff. However, we notice that as the AE chooses to reveal reviews with higher scores the average DM payoff increases and surpasses the average payoff of the DMs who played with the HIGHEST agent. This is an interesting pattern, given that the AE is trained to maximize its own payoff, and its objective does not take the DM's payoff into account.

Finally, Table \ref{table:average_dm_vs_expert_payoffs} presents the average DM and expert payoffs, considering all the experiments (top) and those experiments where the DMs accepted at most 8 hotels. The table demonstrates that the HIGHEST agent yielded the highest average payoffs for both player types, but this is mostly due to a large number of DMs who accepted 9 or 10 hotels. Indeed, when focusing only on DMs who considered the hotels more carefully (bottom table), the average of both the DM and the expert payoffs are quite similar for both agents. The results reflect an interesting property of the HIGHEST agent: It makes many more human DMs accept all (or almost all) of the hotels. This may reflect an interesting difference between human and simulation DMs, to be explored in the future.



\section{Conclusions} 

We consider the problem of automatic expert design for a repeated non-cooperative persuasion game. Our AE is based on the MCTS search algorithm with deep learning models for DM decision and expert's future payoff predictions. Our experiments quantitatively and qualitatively analyse the performance of our AE in comparison to a large variety of alternatives. While our main evaluation is with simulated (automatic) DMs, we also examine the generalizability of our results to experiments with human DMs.

Our work relies on the dataset of \citet{apel2020predicting} for training and testing the various expert models. One limitation of this dataset is its size: It is based on only 10 training and 10 test hotels, each with 7 scored reviews. Moreover, the training set, which is used for training our DMM and VM models, consists of only 408 ten-trial games. We aimed to compensate for this by performing a large number of simulations (1000) for each expert/DM pair, and by reporting 95\% Confidence Intervals (CIs), demonstrating limited overlap between the  95\% CI of our AE and the baselines. Yet, richer datasets in terms of the size and diversity of the hotel sets, as well as the richness of interaction between the human players, are required in order to further validate our results.

In future we would like to extend our AE in three main directions: (a) Designing end-to-end architectures, where the DMM and VM are jointly trained in order to maximize the AE's payoff; (b) Letting the AE generate persuasive language rather than choosing from pre-written reviews; and (c) Considering other AE strategies such as fair payoff division between the expert and the DM, instead of maximizing the AE's payoff.

\section*{Acknowledgments}
We would like to thank the action editor and the reviewers, as well as the members of the IE@Technion NLP group for their valuable feedback and advice. We are also extremely thankful for the valuable guidance and assistance of Christian König-Kersting in conducting the human experiments. This research is partially funded by an ISF personal grant No. 1625/18. The work of R. Apel and M. Tennenholtz is funded by the European Research Council (ERC) under the European Union's Horizon 2020 research and innovation programme (grant agreement No. 740435)

\iftaclpubformat

\else
\fi

\bibliography{tacl2018}
\bibliographystyle{acl_natbib}

\end{document}